\documentclass{article}

\PassOptionsToPackage{numbers, compress}{natbib}
 \usepackage[preprint]{neurips_2026}

\usepackage[utf8]{inputenc} 
\usepackage[T1]{fontenc}    
\usepackage[hidelinks]{hyperref}       
\usepackage{url}            
\usepackage{booktabs}       
\usepackage{amsfonts}       
\usepackage{nicefrac}       
\usepackage{microtype}      
\usepackage{xcolor}         

\usepackage{caption}
\usepackage{multirow}
\usepackage{graphicx}
\usepackage[table, dvipsnames]{xcolor}
\usepackage{amssymb}
\usepackage{pifont}
\newcommand{\cmark}{\ding{51}}%
\newcommand{\xmark}{\ding{55}}%
\usepackage{wrapfig}
\usepackage{tabularx}
\usepackage{booktabs}
\usepackage{colortbl}
\usepackage{array}

\usepackage[capitalize]{cleveref}
\crefname{section}{Sec.}{Secs.}
\Crefname{section}{Section}{Sections}
\Crefname{table}{Table}{Tables}
\crefname{table}{Tab.}{Tabs.}

\title{Knowledge Distillation \\for Visual Autoregressive Models}

\author{
  Elia Peruzzo \\
  Qualcomm AI Research\thanks{Qualcomm AI Research is an initiative of Qualcomm Technologies, Inc.}\\
  \And
  Aritra Bhowmik \\
  Qualcomm AI Research\\
  \And
  Guillaume Sautiere \\
  Qualcomm AI Research\\
  \And
  Yuki M Asano \\
  University of Technology Nuremberg\\
  \And
  Amirhossein Habibian \\
  Qualcomm AI Research\\
}

\begin{document}


\newcommand{\head}[1]{{\smallskip\noindent\textbf{#1}}}
\newcommand{\alert}[1]{{\color{red}{#1}}}
\newcommand{\sm}{\scriptsize}
\newcommand{\eq}[1]{(\ref{eq:#1})}

\newcommand{\Th}[1]{\textsc{#1}}
\newcommand{\mr}[2]{\multirow{#1}{*}{#2}}
\newcommand{\mc}[2]{\multicolumn{#1}{c}{#2}}
\newcommand{\tb}[1]{\textbf{#1}}
\newcommand{\ul}[1]{\underline{#1}}
\newcommand{\ch}{\checkmark}

\newcommand{\red}[1]{{\color{red}{#1}}}
\newcommand{\blue}[1]{{\color{blue}{#1}}}
\newcommand{\green}[1]{\color{green}{#1}}
\newcommand{\gray}[1]{{\color{gray}{#1}}}

\newcommand{\citeme}[1]{\red{[XX]}}
\newcommand{\refme}[1]{\red{(XX)}}

\newcommand{\fig}[2][1]{\includegraphics[width=#1\linewidth]{fig/#2}}
\newcommand{\figh}[2][1]{\includegraphics[height=#1\linewidth]{fig/#2}}


\newcommand{\tran}{^\top}
\newcommand{\mtran}{^{-\top}}
\newcommand{\zcol}{\mathbf{0}}
\newcommand{\zrow}{\zcol\tran}

\newcommand{\ind}{\mathbbm{1}}
\newcommand{\expect}{\mathbb{E}}
\newcommand{\nat}{\mathbb{N}}
\newcommand{\zahl}{\mathbb{Z}}
\newcommand{\real}{\mathbb{R}}
\newcommand{\proj}{\mathbb{P}}
\newcommand{\prob}{\mathbf{Pr}}
\newcommand{\normal}{\mathcal{N}}

\newcommand{\mif}{\textrm{if}\ }
\newcommand{\other}{\textrm{otherwise}}
\newcommand{\minimize}{\textrm{minimize}\ }
\newcommand{\maximize}{\textrm{maximize}\ }
\newcommand{\st}{\textrm{subject\ to}\ }

\newcommand{\id}{\operatorname{id}}
\newcommand{\const}{\operatorname{const}}
\newcommand{\sgn}{\operatorname{sgn}}
\newcommand{\var}{\operatorname{Var}}
\newcommand{\mean}{\operatorname{mean}}
\newcommand{\trace}{\operatorname{tr}}
\newcommand{\diag}{\operatorname{diag}}
\newcommand{\vect}{\operatorname{vec}}
\newcommand{\cov}{\operatorname{cov}}
\newcommand{\sign}{\operatorname{sign}}
\newcommand{\prj}{\operatorname{proj}}

\newcommand{\softmax}{\operatorname{softmax}}
\newcommand{\clip}{\operatorname{clip}}

\newcommand{\defn}{\mathrel{:=}}
\newcommand{\peq}{\mathrel{+\!=}}
\newcommand{\meq}{\mathrel{-\!=}}

\newcommand{\floor}[1]{\left\lfloor{#1}\right\rfloor}
\newcommand{\ceil}[1]{\left\lceil{#1}\right\rceil}
\newcommand{\inner}[1]{\left\langle{#1}\right\rangle}
\newcommand{\norm}[1]{\left\|{#1}\right\|}
\newcommand{\abs}[1]{\left|{#1}\right|}
\newcommand{\frob}[1]{\norm{#1}_F}
\newcommand{\card}[1]{\left|{#1}\right|\xspace}
\newcommand{\divg}[2]{{#1\ ||\ #2}}
\newcommand{\diff}{\mathrm{d}}
\newcommand{\der}[3][]{\frac{d^{#1}#2}{d#3^{#1}}}
\newcommand{\pder}[3][]{\frac{\partial^{#1}{#2}}{\partial{#3^{#1}}}}
\newcommand{\ipder}[3][]{\partial^{#1}{#2}/\partial{#3^{#1}}}
\newcommand{\dder}[3]{\frac{\partial^2{#1}}{\partial{#2}\partial{#3}}}

\newcommand{\wb}[1]{\overline{#1}}
\newcommand{\wt}[1]{\widetilde{#1}}

\def\xssp{\hspace{1pt}}
\def\ssp{\hspace{3pt}}
\def\msp{\hspace{5pt}}
\def\lsp{\hspace{12pt}}

\newcommand{\cA}{\mathcal{A}}
\newcommand{\cB}{\mathcal{B}}
\newcommand{\cC}{\mathcal{C}}
\newcommand{\cD}{\mathcal{D}}
\newcommand{\cE}{\mathcal{E}}
\newcommand{\cF}{\mathcal{F}}
\newcommand{\cG}{\mathcal{G}}
\newcommand{\cH}{\mathcal{H}}
\newcommand{\cI}{\mathcal{I}}
\newcommand{\cJ}{\mathcal{J}}
\newcommand{\cK}{\mathcal{K}}
\newcommand{\cL}{\mathcal{L}}
\newcommand{\cM}{\mathcal{M}}
\newcommand{\cN}{\mathcal{N}}
\newcommand{\cO}{\mathcal{O}}
\newcommand{\cP}{\mathcal{P}}
\newcommand{\cQ}{\mathcal{Q}}
\newcommand{\cR}{\mathcal{R}}
\newcommand{\cS}{\mathcal{S}}
\newcommand{\cT}{\mathcal{T}}
\newcommand{\cU}{\mathcal{U}}
\newcommand{\cV}{\mathcal{V}}
\newcommand{\cW}{\mathcal{W}}
\newcommand{\cX}{\mathcal{X}}
\newcommand{\cY}{\mathcal{Y}}
\newcommand{\cZ}{\mathcal{Z}}

\newcommand{\vA}{\mathbf{A}}
\newcommand{\vB}{\mathbf{B}}
\newcommand{\vC}{\mathbf{C}}
\newcommand{\vD}{\mathbf{D}}
\newcommand{\vE}{\mathbf{E}}
\newcommand{\vF}{\mathbf{F}}
\newcommand{\vG}{\mathbf{G}}
\newcommand{\vH}{\mathbf{H}}
\newcommand{\vI}{\mathbf{I}}
\newcommand{\vJ}{\mathbf{J}}
\newcommand{\vK}{\mathbf{K}}
\newcommand{\vL}{\mathbf{L}}
\newcommand{\vM}{\mathbf{M}}
\newcommand{\vN}{\mathbf{N}}
\newcommand{\vO}{\mathbf{O}}
\newcommand{\vP}{\mathbf{P}}
\newcommand{\vQ}{\mathbf{Q}}
\newcommand{\vR}{\mathbf{R}}
\newcommand{\vS}{\mathbf{S}}
\newcommand{\vT}{\mathbf{T}}
\newcommand{\vU}{\mathbf{U}}
\newcommand{\vV}{\mathbf{V}}
\newcommand{\vW}{\mathbf{W}}
\newcommand{\vX}{\mathbf{X}}
\newcommand{\vY}{\mathbf{Y}}
\newcommand{\vZ}{\mathbf{Z}}

\newcommand{\va}{\mathbf{a}}
\newcommand{\vb}{\mathbf{b}}
\newcommand{\vc}{\mathbf{c}}
\newcommand{\vd}{\mathbf{d}}
\newcommand{\ve}{\mathbf{e}}
\newcommand{\vf}{\mathbf{f}}
\newcommand{\vg}{\mathbf{g}}
\newcommand{\vh}{\mathbf{h}}
\newcommand{\vi}{\mathbf{i}}
\newcommand{\vj}{\mathbf{j}}
\newcommand{\vk}{\mathbf{k}}
\newcommand{\vl}{\mathbf{l}}
\newcommand{\vm}{\mathbf{m}}
\newcommand{\vn}{\mathbf{n}}
\newcommand{\vo}{\mathbf{o}}
\newcommand{\vp}{\mathbf{p}}
\newcommand{\vq}{\mathbf{q}}
\newcommand{\vr}{\mathbf{r}}
\newcommand{\Vs}{\mathbf{s}}
\newcommand{\vt}{\mathbf{t}}
\newcommand{\vu}{\mathbf{u}}
\newcommand{\vv}{\mathbf{v}}
\newcommand{\vw}{\mathbf{w}}
\newcommand{\vx}{\mathbf{x}}
\newcommand{\vy}{\mathbf{y}}
\newcommand{\vz}{\mathbf{z}}

\newcommand{\vone}{\mathbf{1}}
\newcommand{\vzero}{\mathbf{0}}

\newcommand{\valpha}{{\boldsymbol{\alpha}}}
\newcommand{\vbeta}{{\boldsymbol{\beta}}}
\newcommand{\vgamma}{{\boldsymbol{\gamma}}}
\newcommand{\vdelta}{{\boldsymbol{\delta}}}
\newcommand{\vepsilon}{{\boldsymbol{\epsilon}}}
\newcommand{\vzeta}{{\boldsymbol{\zeta}}}
\newcommand{\veta}{{\boldsymbol{\eta}}}
\newcommand{\vtheta}{{\boldsymbol{\theta}}}
\newcommand{\viota}{{\boldsymbol{\iota}}}
\newcommand{\vkappa}{{\boldsymbol{\kappa}}}
\newcommand{\vlambda}{{\boldsymbol{\lambda}}}
\newcommand{\vmu}{{\boldsymbol{\mu}}}
\newcommand{\vnu}{{\boldsymbol{\nu}}}
\newcommand{\vxi}{{\boldsymbol{\xi}}}
\newcommand{\vomikron}{{\boldsymbol{\omikron}}}
\newcommand{\vpi}{{\boldsymbol{\pi}}}
\newcommand{\vrho}{{\boldsymbol{\rho}}}
\newcommand{\vsigma}{{\boldsymbol{\sigma}}}
\newcommand{\vtau}{{\boldsymbol{\tau}}}
\newcommand{\vupsilon}{{\boldsymbol{\upsilon}}}
\newcommand{\vphi}{{\boldsymbol{\phi}}}
\newcommand{\vchi}{{\boldsymbol{\chi}}}
\newcommand{\vpsi}{{\boldsymbol{\psi}}}
\newcommand{\vomega}{{\boldsymbol{\omega}}}

\newcommand{\rLambda}{\mathrm{\Lambda}}
\newcommand{\rSigma}{\mathrm{\Sigma}}

\newcommand{\vLambda}{\bm{\rLambda}}
\newcommand{\vSigma}{\bm{\rSigma}}


\makeatletter
\newcommand{\vast}[1]{\bBigg@{#1}}
\makeatother

\makeatletter
\newcommand*\bdot{\mathpalette\bdot@{.7}}
\newcommand*\bdot@[2]{\mathbin{\vcenter{\hbox{\scalebox{#2}{$\m@th#1\bullet$}}}}}
\makeatother

\makeatletter
\DeclareRobustCommand\onedot{\futurelet\@let@token\@onedot}
\def\@onedot{\ifx\@let@token.\else.\null\fi\xspace}

\def\eg{\emph{e.g}\onedot} \def\Eg{\emph{E.g}\onedot}
\def\ie{\emph{i.e}\onedot} \def\Ie{\emph{I.e}\onedot}
\def\cf{\emph{cf}\onedot} \def\Cf{\emph{Cf}\onedot}
\def\etc{\emph{etc}\onedot} \def\vs{\emph{vs}\onedot}
\def\wrt{w.r.t\onedot} \def\dof{d.o.f\onedot} \def\aka{a.k.a\onedot}
\def\etal{\emph{et al}\onedot}
\makeatother

\newcommand{\methodname}{\textsc{VarKD}}
\newcommand{\best}[1]{\textbf{#1}}
\newcommand{\second}[1]{\underline{#1}}
\newcommand{\arrow}[1]{$\rightarrow$}
\newcommand{\supp}{\emph{Supp.~Mat.}}

\newcommand{\elia}[1]{\textcolor{blue}{Elia: #1}}
\definecolor{mypurple}{HTML}{a38cf4}
\maketitle


\begin{abstract}
Autoregressive (AR) image generation models are highly expressive but computationally intensive, motivating effective model compression. Knowledge distillation (KD) is a natural approach for model compression and has been widely studied in language modeling, yet its behavior in visual AR generation remains underexplored. In this work, we present the first systematic study of distillation strategies for AR image models.
Our analysis shows that while standard distillation can yield meaningful gains, recent methods developed for language do not directly transfer to images: long decoding horizons and visual token ambiguity make teacher supervision unreliable especially under student‑conditioned contexts.
To address this, we propose \methodname{}, a distillation framework for visual autoregressive models that distills on student samples while selectively applying teacher supervision and reducing token-level ambiguity.
Experiments on ImageNet across multiple AR backbones show that \methodname{} consistently outperforms prior distillation baselines, narrowing the gap to large-scale models.
\end{abstract}

\vspace{0.3em}
\begin{center}
\begin{minipage}{0.8\textwidth}
\textbf{Project page:} \url{https://qualcomm-ai-research.github.io/varkd/}
\end{minipage}
\end{center}
\section{Introduction}
\label{sec:introduction}

Autoregressive (AR) models have become a central framework for sequence generation, driving recent advances in large language models and, increasingly, in image generation. Their appeal comes from a simple factorization of the joint distribution into a sequence of conditional predictions, enabling stable maximum likelihood training, strong global coherence, and a natural interface with multimodal and language-conditioned workflows~\cite{vaswani2017attention, radford2019language, van2016conditional}.
In vision, combining this paradigm with discrete tokenization and Transformer decoders has produced scalable AR image generators with competitive fidelity and strong prompt faithfulness~\cite{van2017neural,razavi2019generating,ramesh2021zero,sun2024autoregressive,tian2024visual,tang2024hart,han2025infinity}.

Despite their strong modeling capabilities, AR image models remain inefficient because image synthesis relies on long-horizon next-token decoding, often requiring hundreds or thousands of sequential steps per image~\cite{chen2020generative, lee2022autoregressive, yu2022scaling}. To address this bottleneck, most prior works have focused on accelerating decoding, for example, through speculative, relaxed, or parallel decoding, while leaving the underlying autoregressive model unchanged~\cite{jang2024lantern,peruzzo2026multi,he2024zipar}.
Although effective in reducing latency, these approaches implicitly assume access to a large, high-capacity autoregressive model whose robustness is preserved under faster decoding.

Another way to improve efficiency, beyond faster decoding, is through \emph{model-capacity efficiency}: training compact autoregressive image models that retain the robustness and generative quality of their larger counterparts. This is difficult in the visual domain because AR image models operate over much longer horizons than language models~\cite{chen2020generative, lee2022autoregressive, yu2022scaling}, rely on highly ambiguous discrete token vocabularies~\cite{jang2024lantern, wang2025tokenbridge}, and exhibit strong spatial coupling, where local prediction errors can propagate into global semantic artifacts~\cite{ren2025beyond, wang2025visual}. These properties make visual AR models especially sensitive to the training-inference mismatch induced by teacher forcing: during training, models are optimized on data prefixes, but at inference they must condition on their own predictions~\cite{bengio2015scheduled, xu2019rethinking, lamb2016professor}. When model capacity is limited, even small early errors become harder to recover from and can compound over hundreds or thousands of decoding steps, causing compact autoregressive image models to degrade rapidly at inference time~\cite{arora2022exposure}.

\begin{figure}[t]
    \centering
    \includegraphics[width=\linewidth]{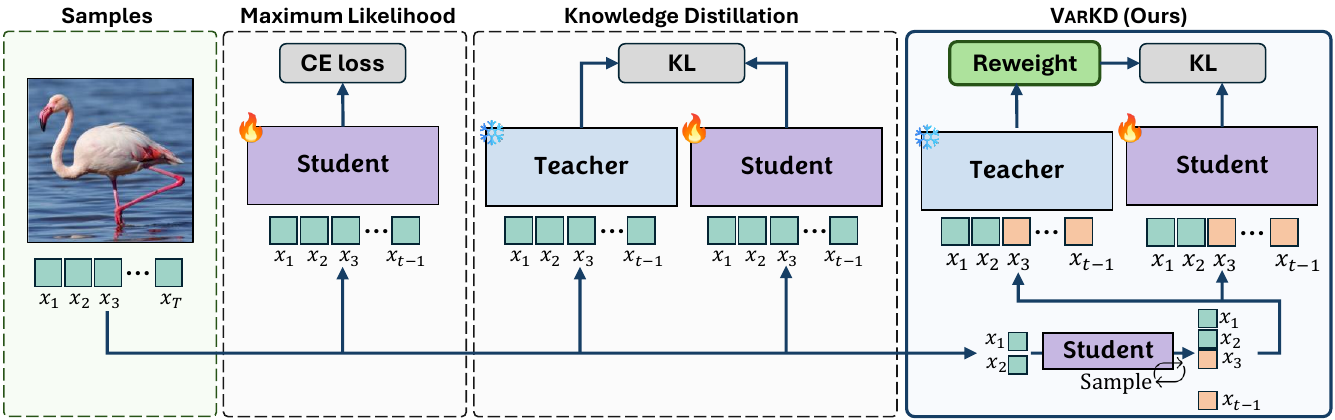}
    \caption{Comparison of training paradigms for autoregressive (AR) image models. We assume access to a dataset of real images (left), represented as sequences of discrete tokens. In standard maximum-likelihood training, the model is trained via teacher forcing, optimizing a cross-entropy loss against ground-truth tokens. In knowledge distillation (KD), a pre-trained teacher guides training by matching the student’s predictive distribution under ground-truth prefixes. In \methodname{} (right), we instead generate samples from the student conditioned on a partial ground-truth context and use the teacher to score these rollouts. We further introduce confidence-based reweighting to improve the reliability of the distillation signal.
    }
    \label{fig:teaser} 
    \vspace{-1em}
\end{figure}

Knowledge distillation (KD) naturally addresses this capacity gap by transferring the behavior of a large teacher to a compact student~\cite{bucilua2006model,hinton2015distilling}. In classification and language modeling, supervised KD~\cite{hinton2015distilling} and sequence-level KD~\cite{kim2016sequence} typically provide supervision under data-distributed inputs or teacher-generated sequences. Recently, the field has shifted toward \emph{on-policy} distillation~\cite{agarwal2024policy, lu2025onpolicydistillation}, which trains students on their own generated sequences to align supervision with inference-time rollouts, thereby reducing the training--inference mismatch and improving the performance. We summarize different training paradigms for AR image models in \Cref{fig:teaser}, highlighting the different data-source and supervision signals.

However, our empirical investigation suggests that directly transferring these strategies to visual autoregressive generation is non-trivial. In particular, on-policy distillation in the visual domain is hampered by image-specific challenges: long decoding horizons and intrinsic token ambiguity cause student samples to rapidly drift off the data manifold~\cite{ren2025beyond,jang2024lantern}, producing contexts where teacher predictions become diffuse and weakly informative. 

Motivated by these findings, we propose \methodname{}, a training framework that provides more reliable supervision for compact visual autoregressive models. \methodname{} distills under student-generated contexts, while improving generation quality at training time by conditioning on a ground-truth prefix. To enhance supervision quality, we apply a reweighting of the loss to down-weight and filter ambiguous, low-confidence teacher predictions, and compute the distillation loss in a compressed visual token space to reduce token-level ambiguity. Finally, we improve training efficiency by employing parallel decoding~\cite{he2024zipar}. All components are used only during training, preserving standard next-token autoregressive decoding at inference.


Our contributions are summarized as follows:
\begin{itemize}
    \item We present \textbf{the first systematic study of knowledge distillation strategies for visual autoregressive (AR) models}. We evaluate supervised KD, sequence-level KD, and on-policy distillation methods adapted from language modeling, and identify image-specific failure modes: long decoding horizons and intrinsic token ambiguity render teacher supervision unreliable under student-conditioned contexts.

    \item Motivated by these findings, we introduce \textbf{\methodname{}, a distillation framework tailored to visual AR models}. \methodname{} improves supervision under mixed student-conditioned rollouts by selectively incorporating teacher feedback based on predictive confidence, and consistently outperforms prior distillation baselines across architectures and model scales.
\end{itemize}

\section{Related Work}
\label{sec:related_work}
\subsection{Autoregressive Models for Image Generation}

Autoregressive (AR) models are among the earliest approaches to image generation, factorizing the joint image distribution into a sequence of conditional predictions optimized via maximum likelihood. Early pixel-space models such as PixelRNN~\cite{van2016pixel} and conditional PixelCNN~\cite{van2016conditional} demonstrated that high-fidelity images could be generated through next-token prediction with stable training and exact likelihood evaluation, establishing autoregressive modeling as a principled alternative to adversarial methods despite high computational cost.

The modern resurgence of AR image generation was enabled by discrete latent representations, most notably VQ-VAE~\cite{van2017neural} and VQ-VAE-2~\cite{razavi2019generating}, which compress images into sequences of visual tokens and substantially reduce generation horizon. This paradigm forms the basis of large-scale text-to-image systems such as DALL·E~\cite{ramesh2021zero} and subsequent autoregressive models including Parti~\cite{yu2022scaling}, VQGAN~\cite{esser2021taming}, HART~\cite{tang2024hart}, LlamaGen~\cite{sun2024autoregressive} and SIMPLEAR~\cite{wang2025simplear}, achieving competitive image quality, semantic coherence, and prompt alignment at scale.

Within this family, autoregressive image models vary in both tokenization and generation structure. Tokenizers range from vector-quantized codebooks~\cite{van2017neural,razavi2019generating} and adversarially trained VQGAN vocabularies~\cite{esser2021taming} to AR-oriented holistic tokenizers~\cite{zheng2025holistic}. Such discrete visual token spaces can have multiple perceptually similar vocabularies, introducing token ambiguity in visual AR modeling~\cite{jang2024lantern}. Generation structures span raster-scan pixel prediction in iGPT~\cite{chen2020generative}, patch-level token generation in LlamaGen~\cite{sun2024autoregressive}, and hierarchical or scale-wise formulations such as HQ-Transformer~\cite{you2022locally}, VAR~\cite{tian2024visual}, M-VAR~\cite{ren2024m}, Switti~\cite{voronov2024switti}, and HART~\cite{tang2024hart}. This paradigm has also been adopted in unified decoder-only multimodal systems such as Emu3~\cite{wang2024emu3}, Chameleon~\cite{team2024chameleon}, and Lumina-mGPT~\cite{liu2024lumina}. Despite this diversity, many token-based AR generators still require long visual rollouts increasing latency and susceptibility to error accumulation~\cite{chen2020generative,lee2022autoregressive,yu2022scaling,ren2025beyond}.

\subsection{Improving Efficiency of Autoregressive Image Models}

A fundamental limitation of autoregressive image generation is the inefficiency of next-token decoding at high resolution. When images are represented as sequences of discrete visual tokens, sequence length grows quadratically with spatial resolution, resulting in hundreds or thousands of sequential decoding steps per sample~\cite{chen2020generative, yu2022scaling}. This behavior is characteristic of large-scale autoregressive image Transformers such as iGPT~\cite{chen2020generative}, LlamaGen~\cite{sun2024autoregressive}, and more recent visual autoregressive models~\cite{he2024zipar}, leading to high inference latency and low throughput. Long decoding horizons further amplify error accumulation, as early prediction errors propagate across subsequent tokens~\cite{bengio2015scheduled, ren2025beyond, arora2022exposure}.

To reduce inference latency, prior works accelerate autoregressive decoding through speculative, draft-based, or parallel generation schemes. Speculative Decoding~\cite{leviathan2023fast} introduced draft-and-verify generation, later extended by multi-head or tree-based methods such as Medusa~\cite{cai2024medusa}, feature-level speculative frameworks like EAGLE~\cite{li2024eagle, li2024eagle2}, and distillation-based draft alignment as in DistillSpec~\cite{zhou2023distillspec}. Extensions to the visual domain explicitly account for properties of visual token sequences: LANTERN~\cite{jang2024lantern} accounts for token ambiguity during verification, while MuLo-SD~\cite{peruzzo2026multi} exploits multi-scale structure and spatial locality. Other methods relax strict raster-order decoding via locality-aware or parallel generation, including ZipAR~\cite{he2024zipar}, LPD~\cite{zhang2025locality}, and ARPG~\cite{li2026autoregressive}. Although effective, these methods target decoding-time efficiency rather than rollout-aware training of compact visual AR models.

A complementary line of work improves efficiency through formulation-level changes, including next-scale or hierarchical AR designs such as VAR~\cite{tian2024visual}, M-VAR~\cite{ren2024m}, Switti~\cite{voronov2024switti}, and HART~\cite{tang2024hart}, as well as training-optimization pipelines such as SimpleAR~\cite{wang2025simplear}. 
These approaches improve the efficiency or scalability of visual generation, but they do not directly address how to train compact visual AR students to remain stable under their own rollouts.

\subsection{Knowledge Distillation for Autoregressive Sequence Models}

Knowledge distillation (KD) is a longstanding approach for compressing large neural networks into smaller models, originally based on logit matching and soft-label supervision~\cite{bucilua2006model, hinton2015distilling}. 
In autoregressive sequence modeling, token-level KD typically supervises the student under teacher-forced contexts by matching teacher and student next-token distributions, often treated as a forward Kullback-Leibler objective~\cite{zhang2023towards, wei2024sentence, gu2024miniplm}. 
Sequence-Level Knowledge Distillation (SeqKD)~\cite{kim2016sequence} extended this paradigm by training students on full sequences generated by the teacher, thereby replacing the original target distribution with a teacher-induced supervision distribution. 
While widely adopted in language modeling, these approaches remain off-policy with respect to the student and therefore do not address the training--inference mismatch associated with exposure bias~\cite{arora2022exposure, ranzato2015sequence}.

Classic imitation-learning work shows that policies trained only on expert states can fail when evaluated on their own induced states, leading to compounding errors~\cite{ross2010efficient,ross2011reduction}. 
Recent autoregressive distillation methods build on this view by supervising students under student-induced contexts: ~\citet{lin2020autoregressive} applied imitation-learning-style KD to language generation, while f-DISTILL~\cite{wen2023f} generalized sequence-level KD with alternative $f$-divergences. 
Generalized Knowledge Distillation (GKD)~\cite{agarwal2024policy} (also called on-policy distillation \cite{lu2025onpolicydistillation}), further trains students on self-generated rollouts using teacher probabilities as dense supervision, with related LLM distillation work exploring on-policy or imitation-learning-based objectives to reduce exposure bias~\cite{gu2024miniplm,pozzi2025mitigating}. 
However, these methods have been developed and validated primarily on text tasks, leaving their behavior in visual AR generation unexplored.
 
Unlike text, visual AR models combine long, spatially coupled rollouts with ambiguous token vocabularies, making student-conditioned contexts prone to error accumulation and teacher feedback less reliable under naive on-policy supervision~\cite{chen2020generative,lee2022autoregressive,ren2025beyond}.
This motivates distillation objectives that adapt on-policy learning to the visual domain while providing reliable teacher supervision.
\section{Method}
\label{sec:method}

\subsection{Distillation for Visual Autoregressive Image Generation}
\label{subsec:distillation_visual_ar}

We study knowledge distillation in the setting of autoregressive (AR) image generation.
Let $x_{1:T}=(x_1,\ldots,x_T)$ denote a tokenized image with tokens $x_t\in\mathcal{V}$, with $\mathcal{V}$ the discrete token vocabulary.
An autoregressive image model defines:
\begin{equation}
p(x_{1:T}) = \prod_{t=1}^{T} p(x_t \mid x_{<t}),
\label{eq:ar_factorization}
\end{equation}
where \(x_{<t} := x_{1:t-1}\) is the prefix context. \\

We assume access to a teacher autoregressive model $p_T$ and a lower-capacity student model $p_S$, both defining next-token distributions over the same vocabulary.
The goal is to train the student to match the teacher's predictive behavior while retaining the student's lower inference cost.

In autoregressive image models, \textbf{Knowledge Distillation} can be expressed as matching the teacher and student next-token distributions under a chosen distribution of conditioning prefixes:
\begin{equation}
\mathcal{L}
=
\mathbb{E}_{x_{<t}\sim p_{\mathrm{src}}}
\left[
\mathcal{D}\!\left(
p_T(\cdot\mid x_{<t})
\,\|\,
p_S(\cdot\mid x_{<t})
\right)
\right],
\label{eq:general_kd}
\end{equation}
where \(p_{\mathrm{src}}\) denotes the source distribution over prefix contexts, and the expectation implicitly averages over token positions.
Here, \(\mathcal{D}\) is a chosen divergence between categorical next-token distributions (e.~g.~ KL or Jensen--Shannon divergence).
Different distillation strategies can be expressed as different choices of data source \(p_{\mathrm{src}}\) and divergence $\mathcal{D}$.

\noindent \textbf{Supervised Knowledge Distillation (KD).} Supervised KD~\cite{hinton2015distilling} applies teacher supervision under prefixes drawn from the data distribution.
Concretely, we set \(p_{\mathrm{src}} = p_{\mathrm{data}}\), so prefixes correspond to ground-truth visual token sequences.
The resulting objective is
\begin{equation}
\mathcal{L}_{\mathrm{KD}}
=
\mathbb{E}_{x_{<t}\sim p_{\mathrm{data}}}
\left[
\mathcal{D}\!\left(
p_T(\cdot \mid x_{<t})
\,\|\, 
p_S(\cdot \mid x_{<t})
\right)
\right].
\label{eq:tf_kd}
\end{equation}
This objective exposes the student only to data-distributed prefixes or contexts during training.
Consequently, prefixes encountered at inference, where the student model conditions on its own predictions, are not represented at training time.

\noindent \textbf{Sequence-level distillation (SeqKD).} SeqKD~\cite{kim2016sequence} replaces ground-truth training sequences with full sequences sampled from the teacher model. Classic SeqKD often uses hard teacher-generated targets; here we express its soft distribution-matching analogue to maintain a unified notation.
Specifically, visual token sequences are first generated as \(x_{1:T} \sim p_T(x_{1:T})\) and the student is trained on prefixes induced by these teacher rollouts. In our formulation, this corresponds to setting \(p_{\mathrm{src}} = p_T\). The resulting objective is
\begin{equation}
\mathcal{L}_\mathrm{SeqKD}
=
\mathbb{E}_{x_{<t}\sim p_T}
\left[
\mathcal{D}\!\left(
p_T(\cdot \mid x_{<t})
\,\|\, 
p_S(\cdot \mid x_{<t})
\right)
\right].
\label{eq:seq_kd}
\end{equation}
Compared to supervised KD, SeqKD changes the supervision distribution by replacing data prefixes with teacher-generated ones, but supervision remains off-policy with respect to the student.

\noindent \textbf{Generalized Knowledge Distillation (GKD).} GKD~\cite{agarwal2024policy} applies supervision under contexts generated by the student model itself, often referred to as \emph{on-policy distillation}~\cite{lu2025onpolicydistillation}.
Formally, we set \(p_{\mathrm{src}} = p_S\), so prefixes are induced by student rollouts.
This exposes the student to its own rollout distribution during training.
The corresponding objective is
\begin{equation}
\mathcal{L}_\mathrm{GKD}
=
\mathbb{E}_{x_{<t}\sim p_S}
\left[
\mathcal{D}\!\left(
p_T(\cdot \mid x_{<t})
\,\|\, 
p_S(\cdot \mid x_{<t})
\right)
\right].
\label{eq:on_policy_kd}
\end{equation}

By training on student-generated samples, on-policy distillation reduces the training-inference mismatch.  However, we find that directly applying it to visual AR models yields smaller gains than those observed in language modeling. 

The next section analyzes the underlying causes and introduces practical modifications that make on-policy distillation more stable and effective for visual autoregressive generation.

\subsection{Reliable teacher feedback under student-conditioned contexts}
\label{subsec:student_conditioned_var}

\paragraph{Mixed data--student context distributions.}
A direct application of \cref{eq:on_policy_kd} requires fully student-generated prefixes \(x_{<t}\sim p_S\).
For visual autoregressive models, such free-running rollouts can drift away from data-like contexts, especially during early training or for capacity-limited students.
To regulate the distribution of encountered contexts while retaining exposure to inference-time behavior, we introduce mixed data--student rollouts.

Given a prefix length \(k\), we define a mixed rollout distribution as:
\begin{equation}
p_{\mathrm{mix}}^{k}(x_{1:T})
=
p_{\mathrm{data}}(x_{1:k})
\prod_{t=k+1}^{T} p_S(x_t \mid x_{<t}),
\label{eq:pmix}
\end{equation}
where \(p_{\mathrm{data}}(x_{1:k})\) denotes the data distribution over length \(k\) prefixes.
Under this distribution, the first \(k\) tokens are anchored to ground-truth visual tokens, while the remaining suffix is generated autoregressively by the student.
As a rollout distribution, \(p_{\mathrm{mix}}^{k}\) interpolates between fully data-conditioned sequences (\(k=T\)) and fully student-generated sequences (\(k=0\)).

Unless stated otherwise, we apply the distillation objective under \(p_{\mathrm{mix}}^{k}\), which empirically improves supervision quality without altering the inference-time decoding procedure.

\paragraph{Student-sampled, teacher-scored objective.}
Let \(x_{1:T} \sim p_{\mathrm{mix}}^{k}\) (see \cref{eq:pmix}). For suffix positions \(t > k\), corresponding to tokens generated by the student under the mixed context distribution, we apply distillation via a reverse-KL divergence:

\begin{equation}
\mathcal{L}_{\mathrm{GKD}}
=
\mathbb{E}_{x_{1:T}\sim p_{\mathrm{mix}}^{k},\, t>k}
\left[
\mathcal{D}\!\left(
p_S(\cdot \mid x_{<t})
\,\|\,
p_T(\cdot \mid x_{<t})
\right)
\right].
\label{eq:gkd_mixed}
\end{equation}

This corresponds to on-policy distillation under controlled student rollouts, with supervision applied only to student-generated suffix positions. While \cref{eq:gkd_mixed} mitigates severe distribution shift, teacher predictions on student-generated visual contexts can remain ambiguous, so uniform supervision may still introduce noisy feedback.
We address this through selective teacher supervision.

\paragraph{Selective teacher supervision via entropy reweighting.}
Visual autoregressive models operate over ambiguous discrete token spaces, where teacher uncertainty varies substantially across contexts.
We therefore use teacher predictive entropy to estimate the reliability of each distillation target and down-weight uncertain supervision to prevent them from dominating the training signal.
For a given context \(x_{<t}\), the teacher entropy is:
\begin{equation}
\mathcal{H}\!\left(p_T(\cdot \mid x_{<t})\right)
=
-
\sum_{v\in\mathcal{V}}
p_T(v \mid x_{<t})
\log p_T(v \mid x_{<t}).
\label{eq:teacher_entropy}
\end{equation}
We define teacher confidence as the normalized inverse entropy:
\begin{equation}
\alpha_{x_{<t}}
=
1 -
\frac{
\mathcal{H}\!\left(p_T(\cdot \mid x_{<t})\right)
}{
\log (|\mathcal{V}|)
}, 
\label{eq:teacher_conf}
\end{equation}

where $\log(|\mathcal{V}|)$ is the maximum entropy, corresponding to a uniform distribution over the vocabulary. By construction, \(\alpha_{x_{<t}} \in [0,1]\): low-entropy (i.e., more confident) teacher predictions receive higher weights, while high-entropy predictions are down-weighted.

While teacher entropy is a useful proxy for model confidence, it can become unreliable once the context drifts off the data manifold: the model may remain confident even when the prefix is incorrect. Inspired by speculative decoding \cite{leviathan2023fast}, we therefore introduce a truncation strategy that stops applying the distillation loss once confidence drops below a threshold. In practice, given a time step $\bar{t}$ such that $\alpha_{x_{<\bar{t}}} < \tau$, we set $\alpha_{x_{<t}} = 0$ for all $t > \bar{t}$, effectively ignoring the remainder of the sequence. This truncation prevents increasingly unreliable student-conditioned contexts from contributing to the distillation loss.

Using this confidence score, we reweight the distillation objective as:
\begin{equation}
\mathcal{L}_{\mathrm{VarKD}}
=
\mathbb{E}_{x_{1:T}\sim p_{\mathrm{mix}}^{(k)},\,t>k}
\left[
\alpha_{x_{<t}}
\mathcal{D}\!\left(
p_S(\cdot \mid x_{<t})
\,\|\,
p_T(\cdot \mid x_{<t})
\right)
\right].
\label{eq:weighted_loss}
\end{equation}

\paragraph{Codebook relaxation via compressed-space distillation.}
Discrete visual vocabularies often contain visually similar tokens, so teacher probability mass may be split across multiple tokens that correspond to similar image content.
To reduce this token-level ambiguity, we introduce \emph{codebook relaxation}: a compressed-space distillation loss that groups visually similar tokens during supervision.

Let \(f: \mathcal{V}\rightarrow\tilde{\mathcal{V}}\) map the original vocabulary to a smaller set of token groups obtained by clustering visual token embeddings.
For each compressed token \(\tilde{x}\in\tilde{\mathcal{V}}\), we aggregate the teacher and student distributions as:
\begin{equation}
\tilde{p}_T(\tilde{x} \mid x_{<t})
=
\sum_{x \in f^{-1}(\tilde{x})}
p_T(x \mid x_{<t}),
\qquad
\tilde{p}_S(\tilde{x} \mid x_{<t})
=
\sum_{x \in f^{-1}(\tilde{x})}
p_S(x \mid x_{<t}).
\label{eq:compressed_distributions}
\end{equation}
The compressed-space objective is:

\begin{equation}
\mathcal{L}_{\mathrm{VarKD}}
=
\mathbb{E}_{x_{1:T}\sim p_{\mathrm{mix}}^{(k)},\,t>k}
\left[
\alpha_{x_{<t}}\,
\mathcal{D}\!\left(
\tilde{p}_S(\cdot \mid x_{<t})
\,\|\,
\tilde{p}_T(\cdot \mid x_{<t})
\right)
\right].
\label{eq:compressed_loss}
\end{equation}
This changes only the supervision signal, the student still predicts over the original vocabulary \(\mathcal{V}\) and inference uses the same decoding process.
By grouping visually similar tokens during supervision, compressed-space distillation reduces sensitivity to token-level ambiguity under long student-conditioned rollouts.


\paragraph{Parallel decoding for efficient training.}
Constructing student-conditioned rollouts under \(p_{\mathrm{mix}}^{k}\) requires autoregressive sampling over long suffixes, which can dominate training cost. To mitigate this overhead, we employ parallel decoding~\cite{he2024zipar} during rollout generation. 

Given a partially decoded visual token grid, parallel decoding identifies a \emph{set} of positions \(\mathcal{S}\) that can be predicted simultaneously, and generates all corresponding tokens in a single forward pass:
\begin{equation}
\{x_i : i \in \mathcal{S}\}
\leftarrow
\mathrm{ParallelDecode}(p_S, x_{\mathrm{obs}}),
\end{equation}
where \(x_{\mathrm{obs}}\) denotes the current observed context available to the student.

This procedure is used solely to accelerate rollout construction during training. It does not modify the distillation objective, the model architecture, or the standard autoregressive decoding used at inference time.

\section{Experiments}
\label{sec:experiments}
We evaluate different distillation strategies for visual autoregressive (AR) image generation across multiple teacher--student configurations. Our experiments focus on two representative backbones: LlamaGen~\cite{sun2024autoregressive}, a standard next-token AR Transformer, and ARPG~\cite{li2026autoregressive}, a recent AR model designed to support randomized parallel decoding. We first describe the experimental setup and baselines, then report quantitative and qualitative results, and finally summarize ablations that validate our choices.

\subsection{Setting}
\textbf{Dataset and task} We conduct class-conditional image generation experiments on ImageNet~\cite{russakovsky2015imagenet}. All models are trained following the original training code released with each backbone
, with distillation objectives replacing (or augmenting) the standard next-token cross-entropy loss. We report FID, Inception Score (IS), and Precision/Recall on the ImageNet validation set. Following the Llamagen evaluation protocol, we generate 50k samples per model and compute all metrics on these samples.

\paragraph{Baselines}
We compare against three distillation baselines adapted from language to visual AR generation:
(i) Knowledge Distillation \textbf{(KD)}~\cite{hinton2015distilling}, using token-level logit matching under teacher-forced samples by minimizing the forward KL divergence
(ii) Sequence-level KD \textbf{(SeqKD)}~\cite{kim2016sequence}, which trains on sequences sampled from the teacher (i.e., replacing ground-truth training sequences with teacher-generated samples) while keeping the same token-level distillation objective.
(iii) Generalized KD \textbf{(GKD)}~\cite{agarwal2024policy}, which mixes ground-truth samples with student-conditioned generated samples: with probability $\lambda_{\mathrm{GKD}}$ we draw contexts from student and train the model to match the teacher distribution; otherwise we use ground-truth contexts.

\paragraph{Training details and hyperparameters} We train all the methods on the ImageNet dataset, with a batch size of 256 and learning rage of $5\times 10^{-5}$ on 8 A100 GPUs.
First, we train with KD for 250k iterations. As KD is the most data-efficient approach, requiring no sampling from either the teacher or the student, it is also the fastest to train. We therefore use KD as an alignment stage, initializing all other methods from its best checkpoint. For a fair comparison, we continue training the KD baseline for the same number of iterations as the other methods, but observe no further improvement beyond this point (see \cref{subsec:ablations}). 
For SeqKD, we precompute a teacher-generated dataset of 1.5M samples (approximately the size of ImageNet) to amortize teacher inference cost; training then follows the same hyperparameters as KD.

For GKD, we set $\lambda_{\mathrm{GKD}}=0.1$ and optimize a Jensen--Shannon divergence objective with $\beta=0.5$, and train the model for 20k iterations observing no further improvement beyond this point. We apply GKD (and \methodname) sampling from \emph{conditional student} (see \cref{eq:pmix}) rather than fully free-running generation, as we found full samples to be less stable for smaller students. Concretely, we seed each training example with a ground-truth prefix whose length is sampled uniformly from $[0.25T,\,0.75T]$ (where $T$ is the sequence length), and generate the remaining suffix from the student. This corresponds to training under mixed data--student contexts and substantially improves generation (see \cref{fig:conditional_student_supp} in \supp). 

\methodname{} uses the same base setup as GKD, with the following additions:
(i) \emph{Entropy reweighting}: we reweight the per-token loss with a normalized entropy as defined by \cref{eq:teacher_conf} and drop the suffix using threshold $\tau=0.1$.
(ii) \emph{Codebook relaxation}: we cluster token embeddings with $K$-means to compress the vocabulary by a factor of 4, and compute the distillation loss in the compressed space to reduce supervision noise from token ambiguity (see \supp{} for visualization).
(iii) \emph{Parallel decoding for rollout generation}: we generate student suffixes using ZipAR-style blockwise~\cite{he2024zipar} sampling with window size $4$ during training (only to accelerate generation; inference remains standard autoregressive decoding).


\subsection{Main Results}
\begin{table}[t]
\centering
\caption{Comparison to knowledge distillation methods (KD, SeqKD, GKD) adapted for visual autoregressive models. Results are reported on ImageNet across multiple student sizes and architectures (LlamaGen and ARPG). \methodname{} consistently matches or surpasses prior baselines.}
\label{tab:sota}

\setlength{\tabcolsep}{5pt} 

\begin{tabularx}{\textwidth}{l *{4}{>{\centering\arraybackslash}X}}
\toprule
\textbf{Method} &
\textbf{FID} ($\downarrow$) &
\textbf{IS} ($\uparrow$) &
\textbf{Precision} ($\uparrow$) &
\textbf{Recall} ($\uparrow$) \\
\midrule

Llamagen-B & 6.51 & 156.3 & 0.81 & 0.46 \\
\quad+ KD \cite{hinton2015distilling} & 4.92 & 195.6 & 0.84 & 0.45 \\
\quad+ SeqKD \cite{kim2016sequence} & 5.08 & 197.9 & 0.84 & 0.44 \\
\quad+ GKD \cite{agarwal2024policy} & 4.94 & 194.7 & 0.85 & 0.44 \\
\quad+ \cellcolor{mypurple!10}\methodname &
\cellcolor{mypurple!10}\best{4.58} &
\cellcolor{mypurple!10}\best{203.3} &
\cellcolor{mypurple!10}\best{0.85} &
\cellcolor{mypurple!10}\best{0.46} \\

\midrule

Llamagen-L & 3.07 & 156.0 & 0.83 & 0.52 \\
\quad+ KD \cite{hinton2015distilling} & 3.02 & 244.5 & 0.82 & 0.54 \\
\quad+ SeqKD \cite{kim2016sequence} & 3.10 & \best{258.1} & 0.83 & 0.52 \\
\quad+ GKD \cite{agarwal2024policy} & 2.93 & 248.0 & 0.82 & 0.55 \\
\quad+ \cellcolor{mypurple!10}\methodname &
\cellcolor{mypurple!10}\best{2.83} &
\cellcolor{mypurple!10}242.7 &
\cellcolor{mypurple!10}\best{0.83} &
\cellcolor{mypurple!10}\best{0.55} \\

\midrule

ARPG-L & 2.30 & 309.0 & 0.80 & 0.58 \\
\quad+ KD \cite{hinton2015distilling} & 2.21 & 293.1 & 0.80 & 0.58 \\
\quad+ SeqKD \cite{kim2016sequence} & 2.22 & \best{310.2} & 0.80 & 0.58 \\
\quad+ GKD \cite{agarwal2024policy} & 2.19 & 307.8 & \best{0.81} & 0.59 \\
\quad+ \cellcolor{mypurple!10}\methodname &
\cellcolor{mypurple!10}\best{2.15} &
\cellcolor{mypurple!10}301.3 &
\cellcolor{mypurple!10}0.80 &
\cellcolor{mypurple!10}\best{0.59} \\

\bottomrule
\end{tabularx}

\end{table}
\begin{table*}[!t]
    \centering
    \def\arraystretch{1.0}
    \resizebox{\textwidth}{!}{
    \setlength\tabcolsep{0pt}
    \footnotesize
    \renewcommand{\arraystretch}{0.5}
    \begin{tabular}{c@{\hskip 0.5mm}c@{\hskip 0.5mm}c@{\hskip 0.5mm}c@{\hskip 0.5mm}c@{\hskip 0.5mm}c}
         Teacher & Student & + KD \cite{hinton2015distilling} & + SeqKD \cite{kim2016sequence} & + GKD \cite{agarwal2024policy} & + \methodname \\

         \includegraphics[trim=2 0 0 2,clip,width=0.15\columnwidth]{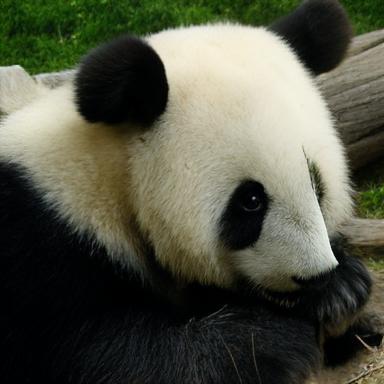} & 
        \includegraphics[trim=2 0 0 2,clip,width=0.15\columnwidth]{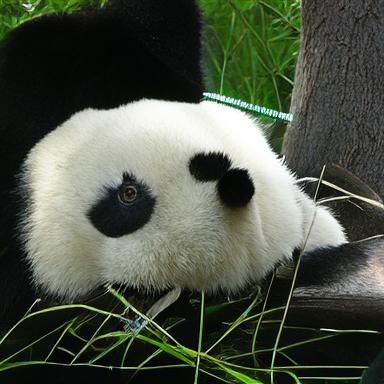} & 
         \includegraphics[trim=2 0 0 2,clip,width=0.15\columnwidth]{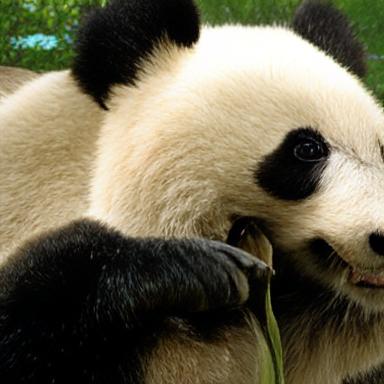} & 
         \includegraphics[trim=2 0 0 2,clip,width=0.15\columnwidth]{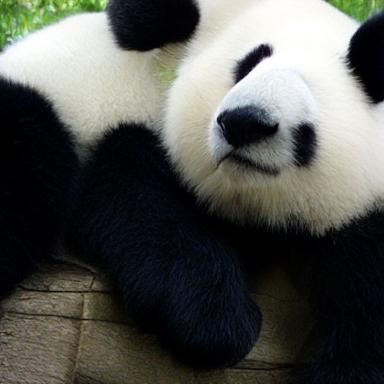} & 
         \includegraphics[trim=2 0 0 2,clip,width=0.15\columnwidth]{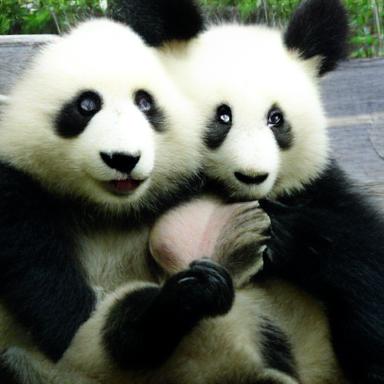} &
         \includegraphics[trim=2 0 0 2,clip,width=0.15\columnwidth]{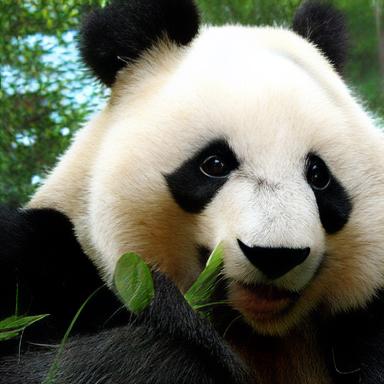} \\

         \includegraphics[trim=2 0 0 2,clip,width=0.15\columnwidth]{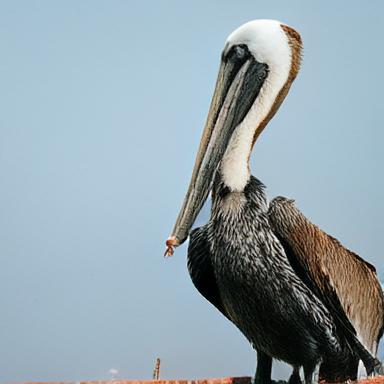} & 
         \includegraphics[trim=2 0 0 2,clip,width=0.15\columnwidth]{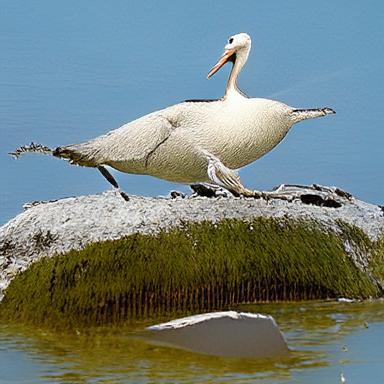} & 
         \includegraphics[trim=2 0 0 2,clip,width=0.15\columnwidth]{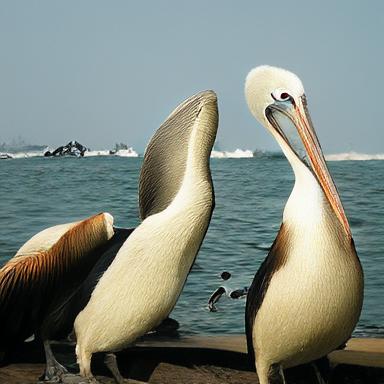} & 
         \includegraphics[trim=2 0 0 2,clip,width=0.15\columnwidth]{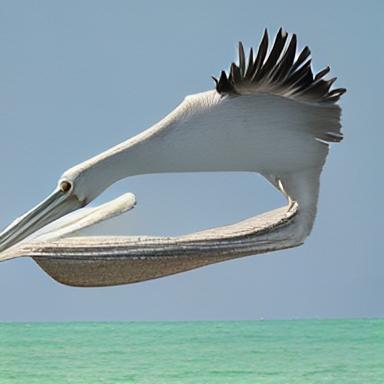} & 
         \includegraphics[trim=2 0 0 2,clip,width=0.15\columnwidth]{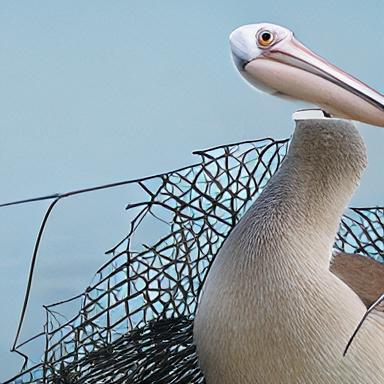} &
         \includegraphics[trim=2 0 0 2,clip,width=0.15\columnwidth]{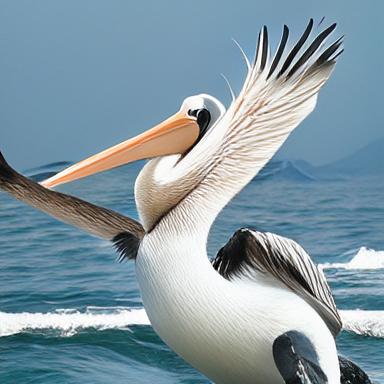} \\

         \includegraphics[trim=2 0 0 2,clip,width=0.15\columnwidth]{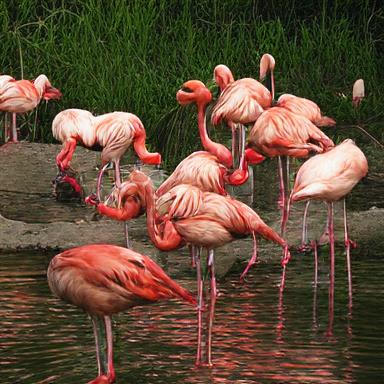} & 
         \includegraphics[trim=2 0 0 2,clip,width=0.15\columnwidth]{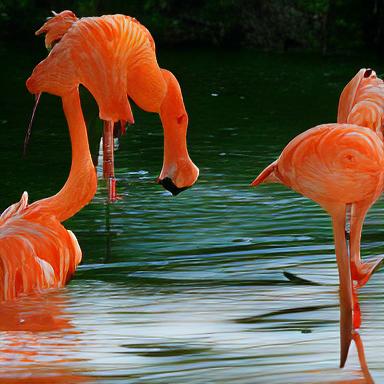} & 
         \includegraphics[trim=2 0 0 2,clip,width=0.15\columnwidth]{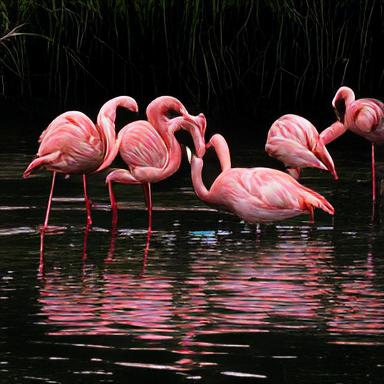} & 
         \includegraphics[trim=2 0 0 2,clip,width=0.15\columnwidth]{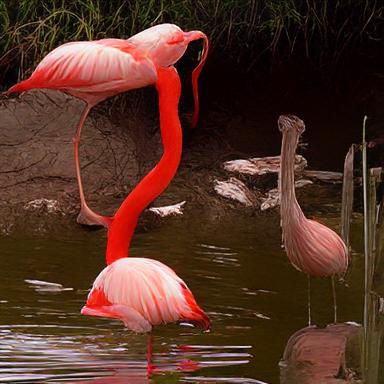} & 
         \includegraphics[trim=2 0 0 2,clip,width=0.15\columnwidth]{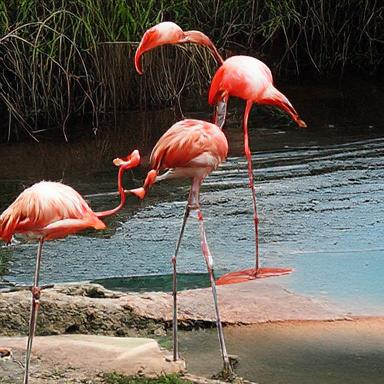} &
         \includegraphics[trim=2 0 0 2,clip,width=0.15\columnwidth]{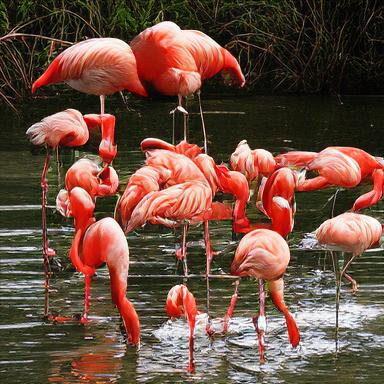} \\
         
    \end{tabular}
    }
    \captionof{figure}{Qualitative comparison showing that \methodname{} reduces spatial artifacts and improves global coherence over prior distillation baselines.}
    \label{fig:qualitative_figure}
    \vspace{-0.3cm}
\end{table*}

\paragraph{Quantitative results.}
\Cref{tab:sota} compares KD, SeqKD, GKD, and \methodname{} across three student architectures (LlamaGen-B, LlamaGen-L, ARPG-L). 
From our empirical evaluation, we observe that: 
First, supervised KD is already a strong baseline for visual AR distillation, recovering a substantial fraction of the teacher--student gap with minimal additional cost compared to on-policy approaches. 
Second, distilling on teacher-generated sequences (SeqKD) is consistently worse than using data-sampled contexts or student-generated rollouts. 
Third, switching to student-generated samples (GKD) yields performance comparable to supervised KD, with more consistent gains for higher-capacity students (e.g., LlamaGen-L and ARPG-L). This supports the intuition that on-policy distillation is most effective when the student is reasonably aligned with the teacher, either through a strong initialization or through objectives that mitigate the noise induced by off-manifold samples. 

Consistent with this view, \methodname{} achieves the best (or tied-best) results across settings and delivers the most reliable improvements in FID. These gains hold across student sizes and across both LlamaGen and ARPG backbones, highlighting the importance of our training-time refinements. Finally, as expected, stronger students are harder to improve; the largest gains are observed for the smaller LlamaGen-B student.

\paragraph{Qualitative results.}
\Cref{fig:qualitative_figure} shows class-conditional samples across methods. Compared to KD and GKD, \methodname{} produces samples with fewer spatial artifacts and improved global coherence. Additional qualitative comparisons are provided in the supplementary material Sec~\ref{app:additional_visual_results}.

\subsection{Ablations}
\label{subsec:ablations}

\paragraph{Method}
\begin{table}[t]
\centering
\begin{minipage}[t]{0.45\textwidth}
\centering
\caption{Method ablation. We analyze the contribution of each component of \methodname.} 
\label{tab:ablation_method}
\resizebox{\columnwidth}{!}{
\begin{tabular}{ccc|cc}
\toprule
\textbf{Entropy} & \textbf{Codebook} & \textbf{Parallel} & \multirow{2}{*}{\textbf{FID ($\downarrow$)}} &  \multirow{2}{*}{\textbf{IS ($\uparrow$)}} \\
\textbf{Rwt.} & \textbf{Rlx.} & \textbf{Decoding}   \\
\midrule
\cellcolor{gray!10}-- & \cellcolor{gray!10}-- &  \cellcolor{gray!10}-- & \cellcolor{gray!10}6.51 & \cellcolor{gray!10}156.3 \\
\xmark & \xmark & \xmark & 4.94 & 194.7 \\
\cmark & \xmark & \xmark & 4.66 & 190.1 \\
\cmark & \cmark & \xmark & 4.60 & 195.7 \\
\cmark & \cmark & \cmark & \textbf{4.58} & \textbf{203.3} \\
\bottomrule
\end{tabular}}

\end{minipage}
\
\begin{minipage}[t]{0.45\textwidth}
\caption{Ablation on teacher size. Effect of teacher capacity on \methodname.}
\label{tab:teacher_size}
\resizebox{\columnwidth}{!}{
\begin{tabular}{lcccc}
\toprule
\textbf{Teacher} $\rightarrow$ \textbf{Student} &\textbf{FID} ($\downarrow$) & \textbf{IS} ($\uparrow$) \\
\midrule
\cellcolor{gray!10}--  & \cellcolor{gray!10}6.51 & \cellcolor{gray!10}156.3  \\
{L \arrow{} B} & \best{4.58} & \best{203.3}  \\
XL \arrow{} B & {4.98} & 198.0 \\
XXL \arrow{} B & {4.82} & 198.2  \\
\bottomrule
\end{tabular}}

\end{minipage}
\vspace{-0.3cm}
\end{table}
We ablate the main design choices of \methodname{} in the LlamaGen-L $\rightarrow$ B in \cref{tab:ablation_method}. The second row corresponds to GKD, where supervision is applied on mixed-prefix, student-conditioned samples. Starting from this setting, we isolate the three components of \methodname{}. Adding teacher-confidence reweighting yields the largest improvement, supporting the need to down-weight unreliable teacher feedback under student-conditioned contexts. Adding codebook relaxation further improves performance, consistent with reduced ambiguity among visually similar tokens and a cleaner supervision signal. Finally, parallel decoding does not degrade quality and even yields slight improvements in generative metrics, possibly due to a regularization effect. Moreover, it accelerates training by $1.3\times$, making it more practical in our setting (see supplementary). 

\paragraph{Effect of Teacher size} We investigate the impact of teacher capacity on student performance in \cref{tab:teacher_size}.
Overall, we observe a trend consistent with prior findings in language-model distillation~\cite{zhong2024revisiting}: a larger teacher does not always produce a better student.
In fact, we find that distillation can improve when using an intermediate or even weaker teacher, and the strongest teacher is not uniformly optimal across tasks or metrics. 
We hypothesize that this behavior arises because very large teachers may produce output distributions that are overly sharp or complex for the student to match, especially under limited student capacity or constrained training budgets.

\paragraph{Two stage approach}

\begin{wrapfigure}{r}{0.48\textwidth}
    \vspace{-1em}
    \centering
    \includegraphics[width=\linewidth]{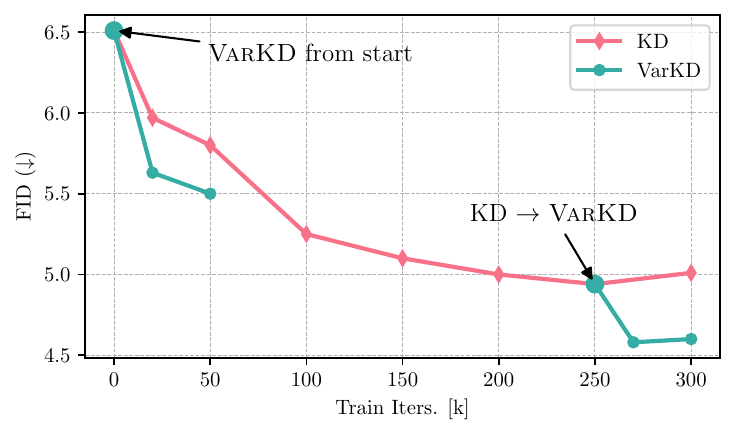}
    \caption{FID vs.~training iterations (lower is better) for KD and \methodname{}. We initialize \methodname{} from two checkpoints, leveraging KD as an efficient alignment stage.}
    \label{fig:placeholder}
    \vspace{-2em}
\end{wrapfigure}

We compare the training efficiency of standard knowledge distillation (KD) with \methodname{}. 
Under the same number of training iterations and starting from the same checkpoint, \methodname{} consistently outperforms KD. 
However, KD operates on a fixed dataset, whereas \methodname{} relies on online sampling, resulting in substantially higher wall-clock cost. 

This motivates a simple two-stage training recipe: (i) first, align the student to the teacher using KD; (ii) then, continue training from this checkpoint with \methodname{}. 
This combination achieves the best overall performance and continues to improve beyond the point where KD alone saturates.

\section{Limitations}
\label{sec:limitations}

A limitation of our study is that we evaluate primarily on class-conditional ImageNet generation, following common AR image-generation benchmarks; extending \methodname{} to open-ended text-to-image settings such as Janus-Pro-style~\cite{chen2025janus} unified generation remains future work. 
In addition, our method relies on teacher confidence and token clustering as proxies for supervision reliability, which may not fully capture semantic correctness or prompt alignment in more complex multimodal generation tasks.

\section{Conclusion}
\label{sec:conclusion}

In this paper, we study knowledge distillation for visual autoregressive image generation and show that language-model distillation methods do not directly address the long horizons, spatial coupling, and token ambiguity of visual AR models.
We present \methodname{}, a rollout-aware distillation framework that improves teacher supervision under mixed data--student contexts through confidence reweighting and compressed-space distillation.
Experiments on ImageNet with LlamaGen and ARPG show that \methodname{} outperforms standard KD, SeqKD, and GKD without changing inference-time decoding.
These results highlight the importance of reliable supervision under student-conditioned rollouts for compressing visual AR models.

\clearpage
\bibliographystyle{plainnat}
\bibliography{refs}

\clearpage
\maketitle

\setcounter{page}{1}

\appendix
\section{Supplementary}
\label{sec:supplementary}

In this supplementary material, we provide additional details and results for \methodname{}.
\cref{app:method} covers discussions on conditional student generation, codebook relaxation, and training-efficiency comparisons, including visual diagnostics for the first two components.
\cref{app:results} reports extended quantitative results across teacher--student configurations and additional qualitative samples.

\subsection{Method Details and Ablations}
\label{app:method}

\paragraph{Conditional Student Generation}
\label{app:conditional_student_generation}
The first modification to standard GKD is the use of mixed data–student contexts (see \cref{eq:pmix}). In practice, we find that seeding student generation with a small number of ground-truth tokens significantly improves both sample quality and the reliability of teacher feedback. We provide visual examples in \Cref{fig:conditional_student_supp}, showing results for both LlamaGen-B and ARPG-L, demonstrating that this strategy generalizes across architectures. Notably, even a short prefix (e.g., 25\% of tokens) yields substantial gains over unconditional generation.

\paragraph{Codebook Relaxation}
\label{app:codebook_relaxation}
\begin{wraptable}{r}{0.6\textwidth}
    \vspace{-2em}
    \centering
    \def\arraystretch{1.0}
    \setlength\tabcolsep{0pt}
    \footnotesize
    \renewcommand{\arraystretch}{0.5}
    \begin{tabular}{ccc}
         $\mathcal{V} = 16384$ & $\mathcal{V'} = 4096$ & $|A - B|$ \\
         \includegraphics[trim=2 0 0 2,clip,width=0.2\columnwidth]{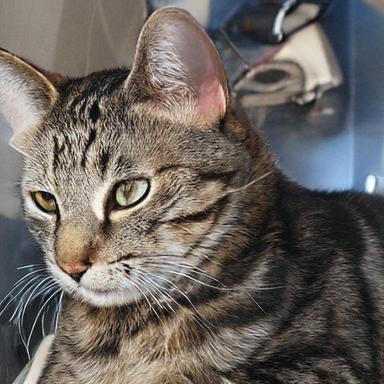} & 
         \includegraphics[trim=2 0 0 2,clip,width=0.2\columnwidth]{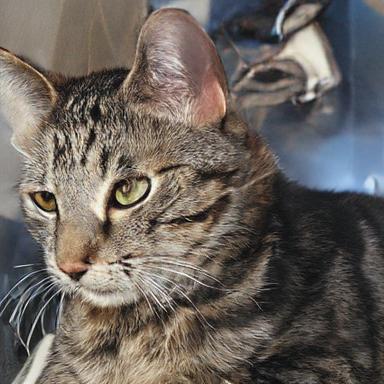} & 
         \includegraphics[trim=2 0 0 2,clip,width=0.2\columnwidth]{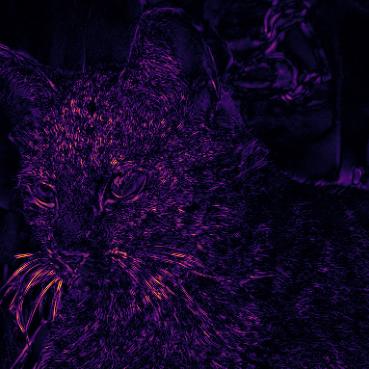} 

    \end{tabular}
    \captionof{figure}{Codebook relaxation. We compare decoding with the full vocabulary (left) and the compressed vocabulary (middle). The visual difference is minimal, as highlighted by the normalized absolute difference (right).}
    \label{fig:codebook_relaxation}
\end{wraptable}
The codebook space of VQ models is often highly redundant, with many tokens corresponding to visually similar output patches. This redundancy can hinder distillation, as the loss treats all token mismatches equally, even when they are nearly indistinguishable in pixel space. 

To address this, we introduce a simple modification. We first perform $k$-means clustering over the codebook embeddings, defining a fixed mapping from the original vocabulary $\mathcal{V}$ to a compressed set $\mathcal{V}' = \mathcal{V} / k$. \Cref{fig:codebook_relaxation} illustrates this process. In practice, we encode an input image and decode it using (a) the full vocabulary and (b) the compressed codebook obtained from the cluster centroids. 

Despite the reduced vocabulary, the visual difference remains minimal. This is confirmed by the normalized absolute difference between the reconstructed images, which shows discrepancies concentrated mainly in high-frequency components.

\paragraph{Training efficiency under different Samples}
\label{app:training_efficiency}
\begin{wraptable}{r}{0.65\textwidth}
\centering
\vspace{-1em}
\caption{Training efficiency comparison across distillation methods. We report throughput (steps/s) and performance under identical settings.}
\begin{tabularx}{0.65\textwidth}{ll|ccc}
\toprule
\textbf{Method} & \textbf{Samples} & \textbf{FID ($\downarrow$)} & \textbf{IS ($\uparrow$)} &\textbf{Steps/s ($\uparrow)$}\\
\midrule
\cellcolor{gray!10}-- & \cellcolor{gray!10}-- &  \cellcolor{gray!10}6.51 & \cellcolor{gray!10}156.3 & \cellcolor{gray!10}-- \\
KD & Data  & 4.92 & 195.6 & 3.50 \\
SeqKD & Teacher & 5.08 & 197.6 & 0.02\textsuperscript{*}\\
SeqKD++ & Cond. Teacher & 4.99 & 204.8 & 0.06 \\
GKD & Cond. Student & 4.94 & 194.7 & 0.15 \\
\methodname & Cond. Student & 4.58 & 203.3 & 0.20 \\
\bottomrule
\end{tabularx}
\end{wraptable}

We further analyze the impact of different data sources on training efficiency. In this section, we focus on comparing the computational cost of distillation methods in terms of steps per second (steps/s), under identical settings of batch size, hardware, and model size. The first row reports the initial student checkpoint.

The second row corresponds to KD, which is the most efficient method since it operates on a fixed external dataset. Notably, it already closes a large portion of the teacher–student gap. SeqKD, on the other hand, trains on teacher-generated data. While its cost can be amortized by caching generated samples, here we report the full online generation setting, which incurs the highest training cost due to the need to sample complete sequences from the teacher.

We also include a conditional-teacher variant (SeqKD++, fourth row), which generates teacher continuations online under mixed-prefix rollouts. Although not standard, this approach provides modest performance gains over SeqKD, at the expense of more complex caching.

Finally, we consider methods that train on student-generated contexts (GKD and \methodname{}). In this regime, we highlight the importance of parallel decoding, which improves throughput by approximately $1.3\times$ without impacting performance (see \cref{tab:ablation_method}).

\subsection{Extended Experiments and Visual Results}
\label{app:results}

\paragraph{Extended teacher--student comparisons.}
\label{app:extended_quantitative_results}
We report a comprehensive evaluation across multiple student–teacher configurations and architectures in \Cref{tab:sota_supp}. For LlamaGen, we consider both B and L variants with teachers of increasing size (L, XL, XXL). For ARPG, we fine-tune the L variant using XL and XXL teachers.

Across architectures, we observe consistent trends. First, standard KD recovers a substantial portion of the teacher–student gap in terms of FID, confirming it as a strong baseline. However, KD is notably sensitive to teacher capacity: performance tends to degrade as the teacher becomes stronger. We hypothesize that this effect arises from the reliance on data-conditioned prefixes, where a stronger teacher induces a sharper and more complex target distribution that is harder for the student to match.

In contrast, this degradation is less pronounced for GKD and \methodname{} (although still present). We attribute this to the different role of the teacher under student-conditioned training: a stronger teacher can provide more informative guidance on off-manifold contexts, helping to correct errors introduced by the student during rollout. 

Overall, the extended results reinforce the conclusions of the main paper: \methodname{} consistently achieves the best or second-best performance across architectures, model sizes, and teacher capacities, demonstrating robust gains in diverse settings.

\paragraph{Additional qualitative comparisons.}
\label{app:additional_visual_results}
We provide additional qualitative samples for LlamaGen-L and ARPG-L in \cref{fig:llamagen_l} and \cref{fig:arpg_samples}.
Across both backbones, \methodname{} produces samples with fewer visible artifacts and better spatial consistency than the distillation baselines.
These examples complement the quantitative results in \Cref{tab:sota_supp}, showing that the gains from \methodname{} are reflected not only in FID but also in visual quality.

\begin{table*}[th]
    \centering
    \def\arraystretch{1.0}
    \resizebox{\textwidth}{!}{
    \setlength\tabcolsep{0pt}
    \footnotesize
    \renewcommand{\arraystretch}{0.5}
    \begin{tabular}{c@{\hskip 0.5mm}c@{\hskip 0.5mm}c@{\hskip 0.5mm}c@{\hskip 0.5mm}c@{\hskip 0.5mm}c}
         Data & Prefix = 0.75\% & Prefix = 0.50\% & Prefix = 0.25\% & Student \\

         \includegraphics[trim=2 0 0 2,clip,width=0.2\columnwidth]{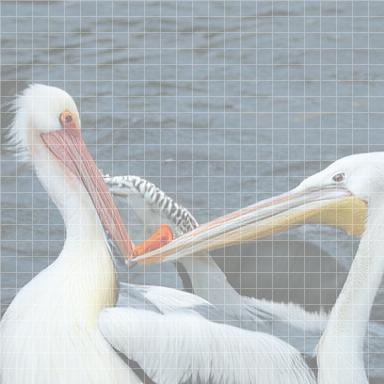} & 
         \includegraphics[trim=2 0 0 2,clip,width=0.2\columnwidth]{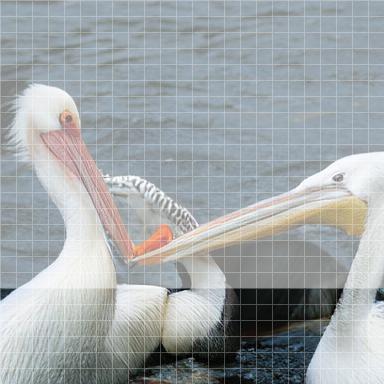} & 
         \includegraphics[trim=2 0 0 2,clip,width=0.2\columnwidth]{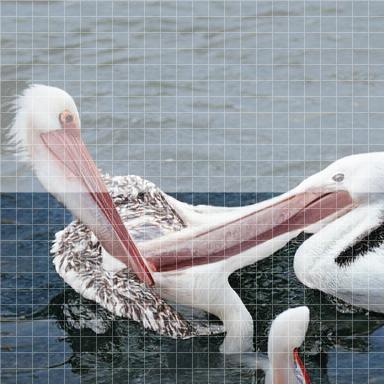} & 
         \includegraphics[trim=2 0 0 2,clip,width=0.2\columnwidth]{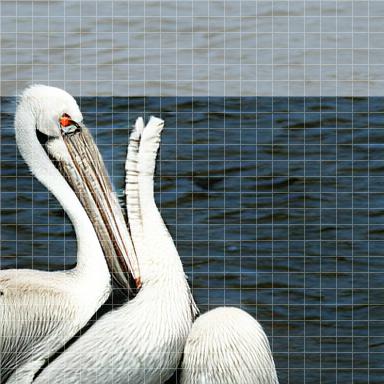} & 
         \includegraphics[trim=2 0 0 2,clip,width=0.2\columnwidth]{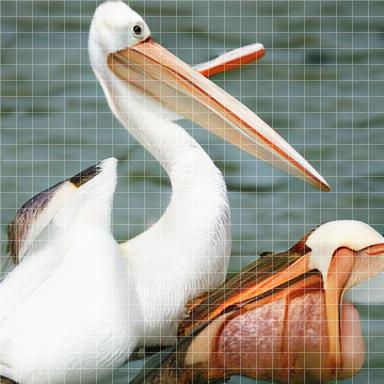}  \\

         \includegraphics[trim=2 0 0 2,clip,width=0.2\columnwidth]{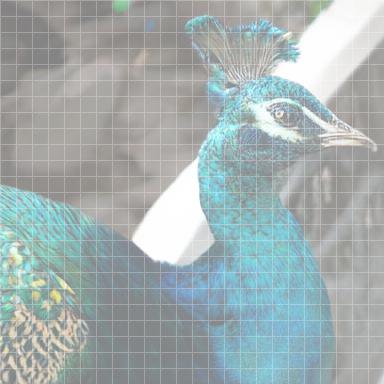} & 
         \includegraphics[trim=2 0 0 2,clip,width=0.2\columnwidth]{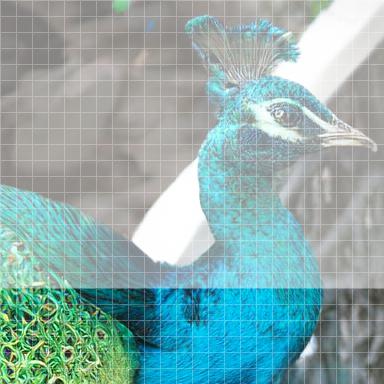} & 
         \includegraphics[trim=2 0 0 2,clip,width=0.2\columnwidth]{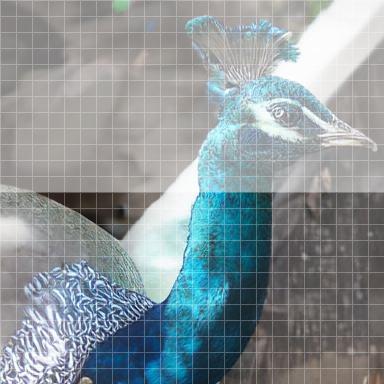} & 
         \includegraphics[trim=2 0 0 2,clip,width=0.2\columnwidth]{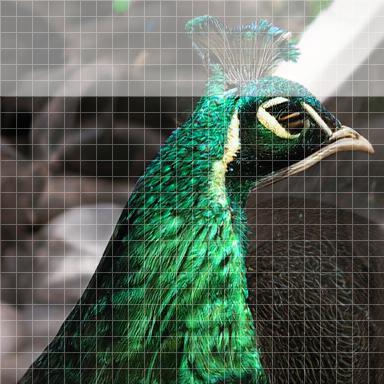} & 
         \includegraphics[trim=2 0 0 2,clip,width=0.2\columnwidth]{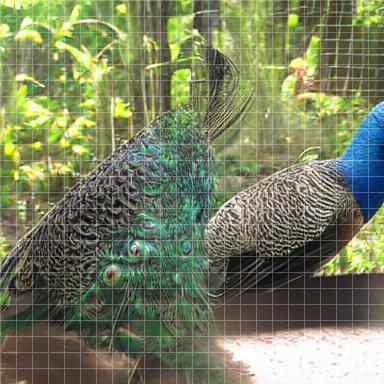} \\

         \includegraphics[trim=2 0 0 2,clip,width=0.2\columnwidth]{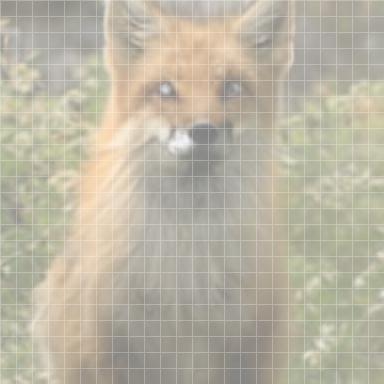} & 
         \includegraphics[trim=2 0 0 2,clip,width=0.2\columnwidth]{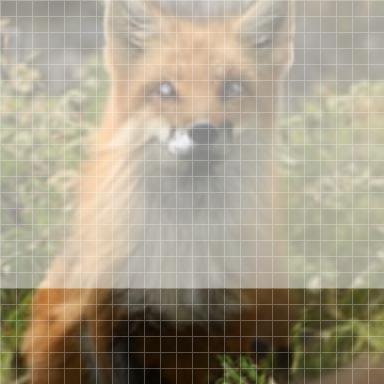} & 
         \includegraphics[trim=2 0 0 2,clip,width=0.2\columnwidth]{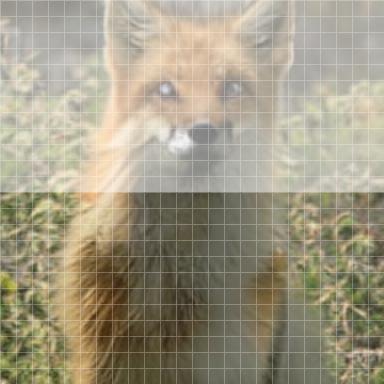} & 
         \includegraphics[trim=2 0 0 2,clip,width=0.2\columnwidth]{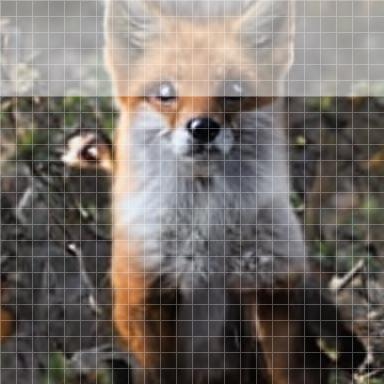} & 
         \includegraphics[trim=2 0 0 2,clip,width=0.2\columnwidth]{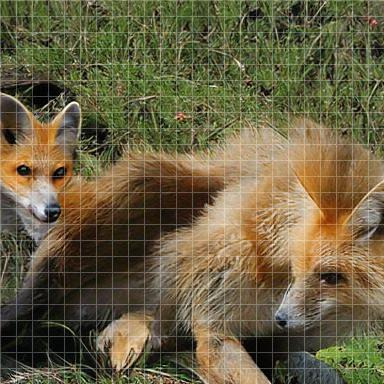} \\
         \multicolumn{5}{c}{(a)} \\
        

         \includegraphics[trim=2 0 0 2,clip,width=0.2\columnwidth]{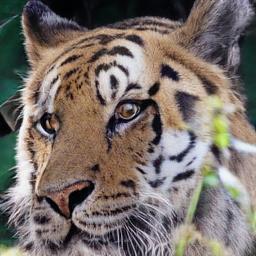} & 
         \includegraphics[trim=2 0 0 2,clip,width=0.2\columnwidth]{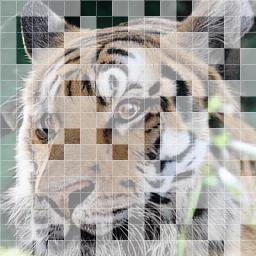} & 
         \includegraphics[trim=2 0 0 2,clip,width=0.2\columnwidth]{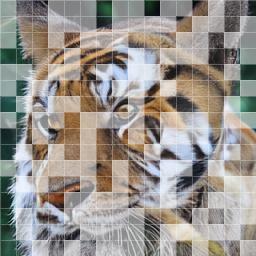} & 
         \includegraphics[trim=2 0 0 2,clip,width=0.2\columnwidth]{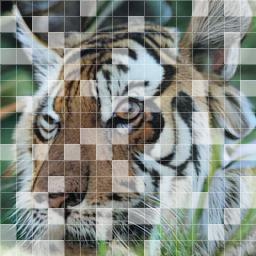} & 
         \includegraphics[trim=2 0 0 2,clip,width=0.2\columnwidth]{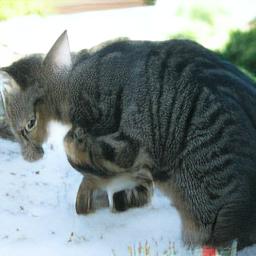} \\

         \includegraphics[trim=2 0 0 2,clip,width=0.2\columnwidth]{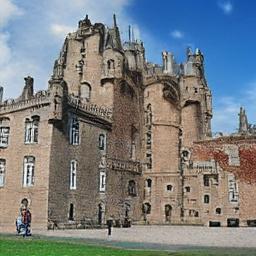} & 
         \includegraphics[trim=2 0 0 2,clip,width=0.2\columnwidth]{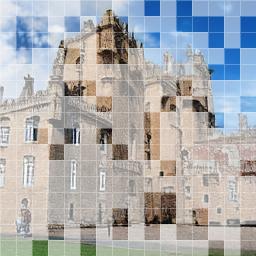} & 
         \includegraphics[trim=2 0 0 2,clip,width=0.2\columnwidth]{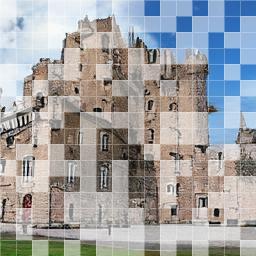} & 
         \includegraphics[trim=2 0 0 2,clip,width=0.2\columnwidth]{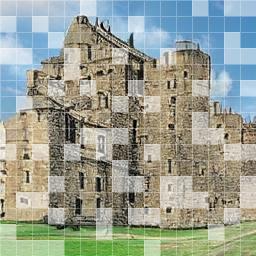} & 
         \includegraphics[trim=2 0 0 2,clip,width=0.2\columnwidth]{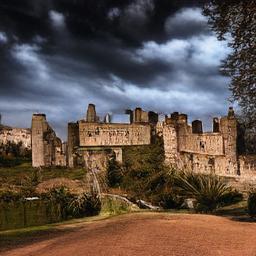} \\

         \includegraphics[trim=2 0 0 2,clip,width=0.2\columnwidth]{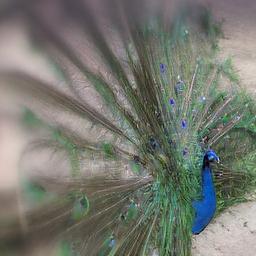} & 
         \includegraphics[trim=2 0 0 2,clip,width=0.2\columnwidth]{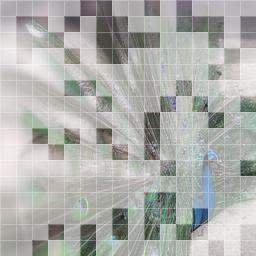} & 
         \includegraphics[trim=2 0 0 2,clip,width=0.2\columnwidth]{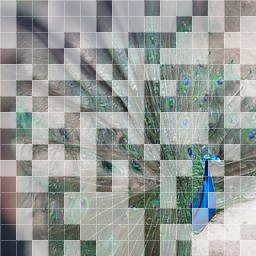} & 
         \includegraphics[trim=2 0 0 2,clip,width=0.2\columnwidth]{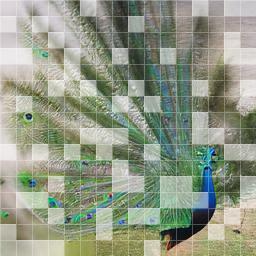} & 
         \includegraphics[trim=2 0 0 2,clip,width=0.2\columnwidth]{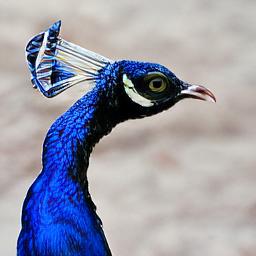} \\
        \multicolumn{5}{c}{(b)} \\

    \end{tabular}
    }
    \captionof{figure}{Conditional student sampling. Prefix tokens from the ground-truth data are highlighted in white. (a) Sampling from LlamaGen-B, which is trained with next-token prediction: the context corresponds to the first \(p\%\) of image tokens. (b) Sampling from ARPG-L, trained with random token order: we randomly mask \(p\%\) of tokens and generate the remaining ones. }
    \label{fig:conditional_student_supp}
\end{table*}
\begin{table}[ht]
\centering
\caption{
Extended teacher--student comparison on the ImageNet validation set.
For each student model, \best{bold} marks the best result across all teacher sizes and distillation methods.
For each fixed teacher--student pair, \second{underlined} marks the best result among the compared distillation methods.
}
\label{tab:sota_supp}

\setlength{\tabcolsep}{5pt} 

\begin{tabularx}{\textwidth}{lXXXXX}
\toprule
\textbf{Teacher} $\rightarrow$ \textbf{Student} & \textbf{Method} &\textbf{FID} ($\downarrow$) & \textbf{IS} ($\uparrow$) & \textbf{Prec.} ($\uparrow$) & \textbf{Recall} ($\uparrow$) \\
\midrule
LlamaGen-B \cite{sun2024autoregressive} & - & 6.51 & 156.3 & 0.81 & 0.46  \\
\cmidrule{2-6}
\multirow{4}{*}{\rotatebox{0}{L \arrow{} B}} & KD \cite{hinton2015distilling} & 4.92 & 195.6 & 0.84 & 0.45 \\
& SeqKD \cite{kim2016sequence} & 5.08 & 197.9 & 0.84 & 0.44 \\
& GKD \cite{agarwal2024policy} & 4.94 & 194.7 & 0.85 & 0.44 \\
& \cellcolor{mypurple!10}\methodname & \cellcolor{mypurple!10}\best{\second{4.58}} & \cellcolor{mypurple!10}\best{\second{203.3}} & \cellcolor{mypurple!10}\best{\second{0.85}} & \cellcolor{mypurple!10}\second{0.46} \\

\cmidrule{2-6}
\multirow{4}{*}{\rotatebox{0}{XL \arrow{} B}} & KD \cite{hinton2015distilling} & 5.35 & 186.5 & 0.82 & 0.46 \\
& SeqKD \cite{kim2016sequence} & 5.37 & \cellcolor{mypurple!10}\second{212.8} & 0.84 & 0.42 \\
& GKD \cite{agarwal2024policy} & 5.02 & 190.1 & 0.82 & \cellcolor{mypurple!10}\second{0.47}  \\
& \cellcolor{mypurple!10}\methodname & \cellcolor{mypurple!10}\second{4.98} & 198.0 & \cellcolor{mypurple!10}\second{0.85} & 0.46 \\

\cmidrule{2-6}
\multirow{4}{*}{\rotatebox{0}{XXL \arrow{} B}} & KD \cite{hinton2015distilling} & 5.72 & 178.7 & 0.83 & 0.44 \\
& SeqKD \cite{kim2016sequence}  &  5.98 & 191.0 & \cellcolor{mypurple!10}\second{0.85} & 0.43 \\
& GKD \cite{agarwal2024policy} & 5.32 & 175.2 & 0.81 & 0.47 \\
& \cellcolor{mypurple!10}\methodname & \cellcolor{mypurple!10}\second{4.82} & \cellcolor{mypurple!10}\second{198.2} & 0.84 & \cellcolor{mypurple!10}\best{\second{0.48}} \\
\midrule

LLamaGen-L \cite{sun2024autoregressive} & - & 3.07 & 156.0 & 0.83 & 0.52  \\
\cmidrule{2-6}

\multirow{4}{*}{\rotatebox{0}{XL \arrow{} L}} & KD \cite{hinton2015distilling} & 3.02 & 244.5 & 0.82 & 0.54 \\
& SeqKD \cite{kim2016sequence} & 3.10 & \cellcolor{mypurple!10}\best{\second{258.1}} & 0.83 & 0.52 \\
& GKD \cite{agarwal2024policy} & 2.93 & 248.0 & 0.82 & 0.55 \\
& \cellcolor{mypurple!10}\methodname & \cellcolor{mypurple!10}\best{\second{2.83}} & 242.7 & \cellcolor{mypurple!10}\best{\second{0.83}} & \cellcolor{mypurple!10}\best{\second{0.55}} \\

\cmidrule{2-6}
\multirow{4}{*}{\rotatebox{0}{XXL \arrow{} L}} & KD \cite{hinton2015distilling} & 3.01 & 246.6 & 0.83 & 0.54 \\
& SeqKD \cite{kim2016sequence} & 3.15 & 238.6 & 0.83 & 0.52 \\
& \cellcolor{mypurple!10}GKD \cite{agarwal2024policy} & \cellcolor{mypurple!10}\second{2.88} & \cellcolor{mypurple!10}\second{248.1} & 0.82 & \cellcolor{mypurple!10}\second{0.55} \\
& \methodname & 2.91 & 246.7 & \cellcolor{mypurple!10}\second{0.83} & 0.54 \\

\midrule
ARPG-L \cite{li2026autoregressive} & - & 2.30 & 309.0 & 0.80 & 0.58  \\
\cmidrule{2-6}
\multirow{4}{*}{\rotatebox{0}{XL \arrow{} L}} & KD \cite{hinton2015distilling} & 2.20 & 296.1 & 0.80 & 0.59 \\
& SeqKD \cite{kim2016sequence} & 2.28 & 305.2 & 0.80 & \cellcolor{mypurple!10}\second{0.59} \\
& GKD \cite{agarwal2024policy} & 2.26 & 310.1 & 0.81 & 0.58 \\
& \cellcolor{mypurple!10}\methodname & \cellcolor{mypurple!10}\second{2.19} & \cellcolor{mypurple!10}\best{\second{313.0}} & \cellcolor{mypurple!10}\best{\second{0.81}} & 0.57 \\
\cmidrule{2-6}

\multirow{4}{*}{\rotatebox{0}{XXL \arrow{} L}} & KD \cite{hinton2015distilling} & 2.21 & 293.1 & 0.80 & 0.58    \\
& SeqKD \cite{kim2016sequence} & 2.22 & \cellcolor{mypurple!10}\second{310.2} & 0.80 & 0.58 \\
& GKD \cite{agarwal2024policy} & 2.19 & 307.8 & \cellcolor{mypurple!10}\second{0.81} & 0.59  \\
& \cellcolor{mypurple!10}\methodname & \cellcolor{mypurple!10}\best{\second{2.15}} & 301.3 & 0.80 & \cellcolor{mypurple!10}\best{\second{0.59}} \\
\bottomrule
\end{tabularx}

\end{table}

\begin{table*}[!t]
    \centering
    \def\arraystretch{1.0}
    \resizebox{\textwidth}{!}{
    \setlength\tabcolsep{0pt}
    \footnotesize
    \renewcommand{\arraystretch}{0.5}
    \begin{tabular}{c@{\hskip 0.5mm}c@{\hskip 0.5mm}c@{\hskip 0.5mm}c@{\hskip 0.5mm}c@{\hskip 0.5mm}c}
         Teacher & Student & + KD \cite{hinton2015distilling} & + SeqKD \cite{kim2016sequence} & + GKD \cite{agarwal2024policy} & + \methodname \\
         
         \includegraphics[trim=2 0 0 2,clip,width=0.15\columnwidth]{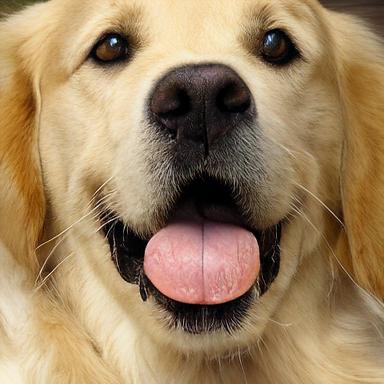} & 
         \includegraphics[trim=2 0 0 2,clip,width=0.15\columnwidth]{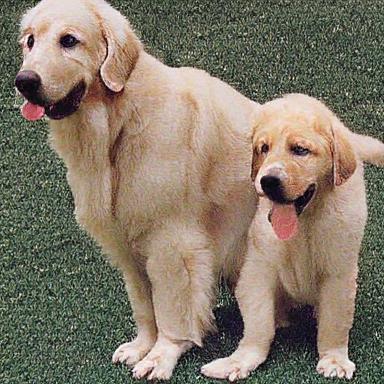} & 
         \includegraphics[trim=2 0 0 2,clip,width=0.15\columnwidth]{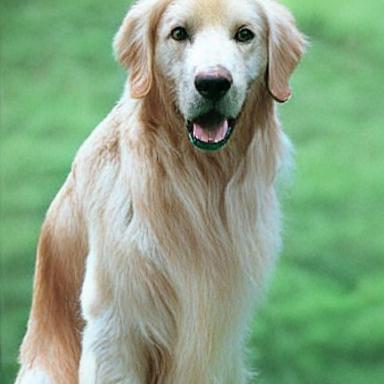} & 
         \includegraphics[trim=2 0 0 2,clip,width=0.15\columnwidth]{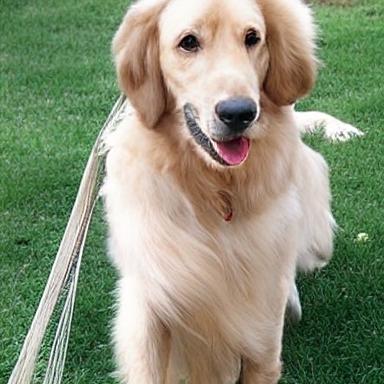} & 
         \includegraphics[trim=2 0 0 2,clip,width=0.15\columnwidth]{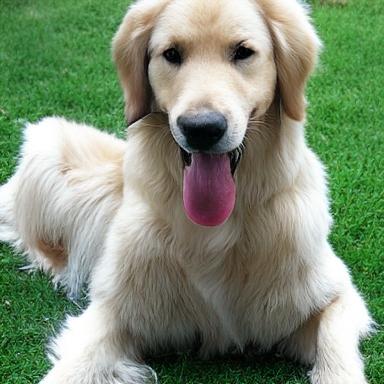} &
         \includegraphics[trim=2 0 0 2,clip,width=0.15\columnwidth]{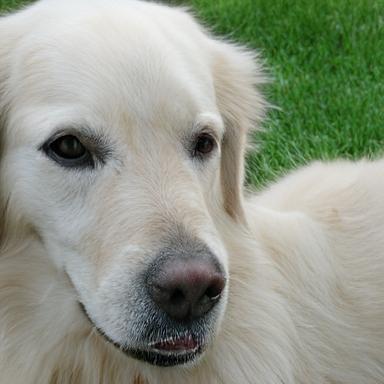} \\

         \includegraphics[trim=2 0 0 2,clip,width=0.15\columnwidth]{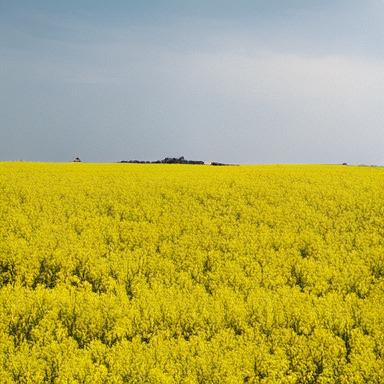} & 
        \includegraphics[trim=2 0 0 2,clip,width=0.15\columnwidth]{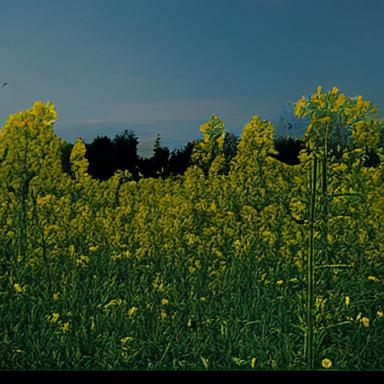} & 
         \includegraphics[trim=2 0 0 2,clip,width=0.15\columnwidth]{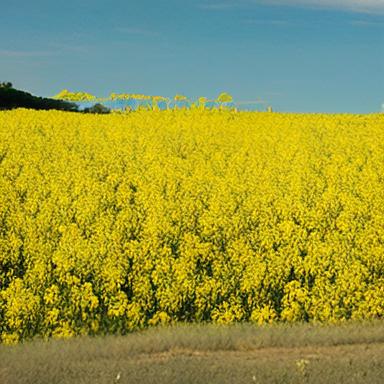} & 
         \includegraphics[trim=2 0 0 2,clip,width=0.15\columnwidth]{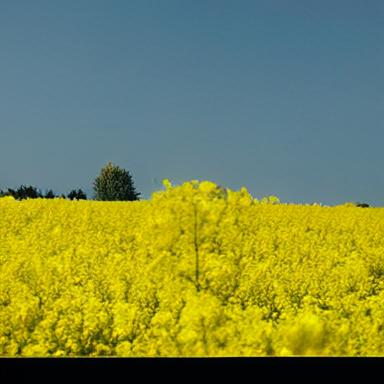} & 
         \includegraphics[trim=2 0 0 2,clip,width=0.15\columnwidth]{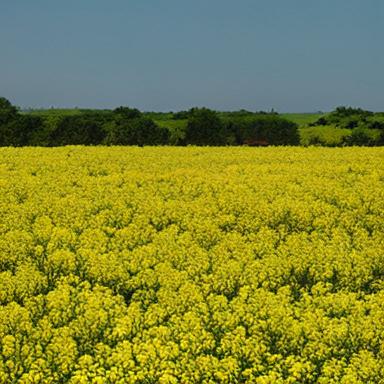} &
         \includegraphics[trim=2 0 0 2,clip,width=0.15\columnwidth]{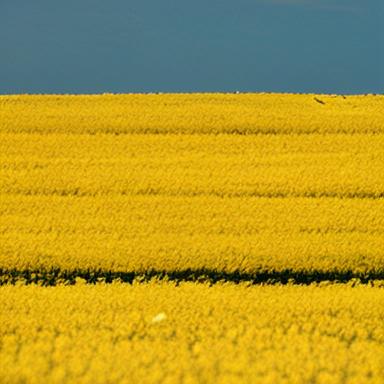} \\

         \includegraphics[trim=2 0 0 2,clip,width=0.15\columnwidth]{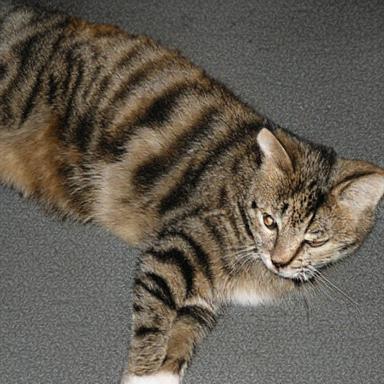} & 
        \includegraphics[trim=2 0 0 2,clip,width=0.15\columnwidth]{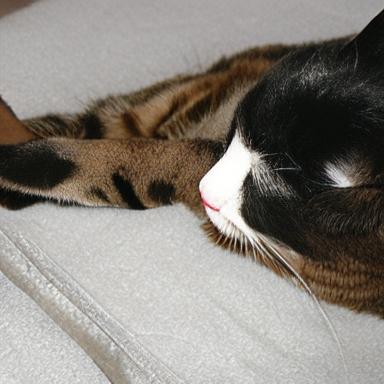} & 
         \includegraphics[trim=2 0 0 2,clip,width=0.15\columnwidth]{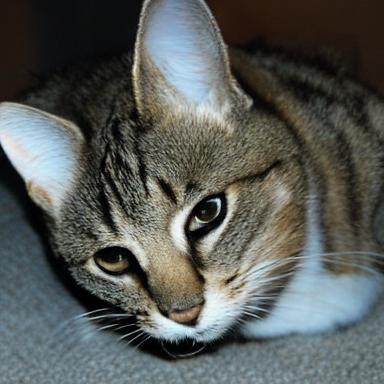} & 
         \includegraphics[trim=2 0 0 2,clip,width=0.15\columnwidth]{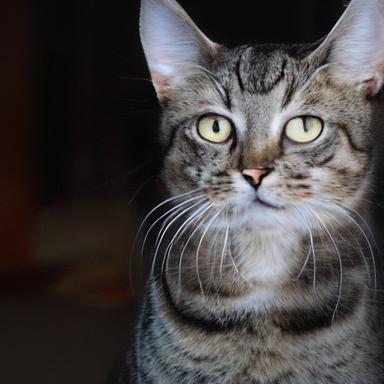} & 
         \includegraphics[trim=2 0 0 2,clip,width=0.15\columnwidth]{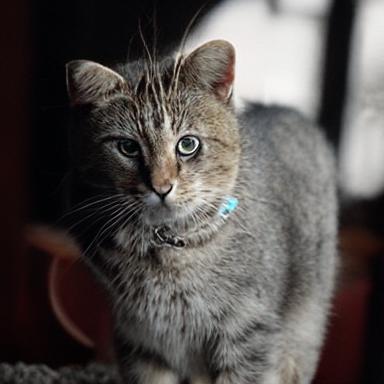} &
         \includegraphics[trim=2 0 0 2,clip,width=0.15\columnwidth]{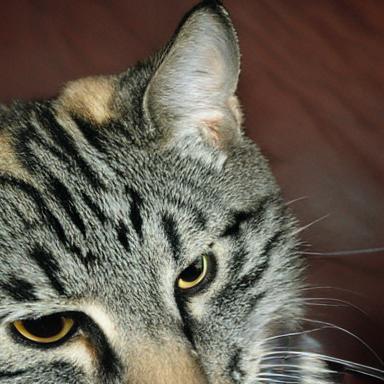} \\

         \includegraphics[trim=2 0 0 2,clip,width=0.15\columnwidth]{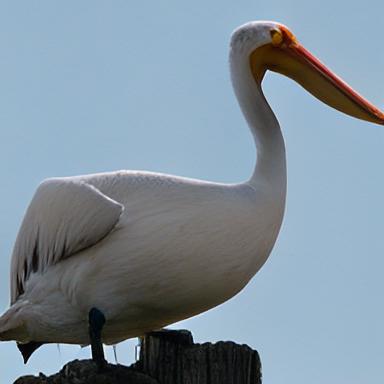} & 
         \includegraphics[trim=2 0 0 2,clip,width=0.15\columnwidth]{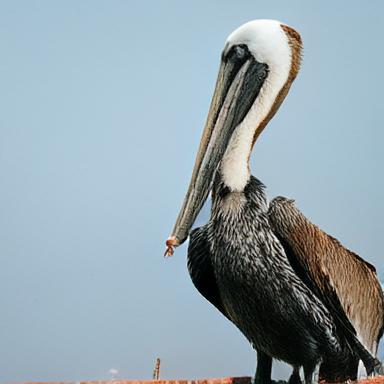} & 
         \includegraphics[trim=2 0 0 2,clip,width=0.15\columnwidth]{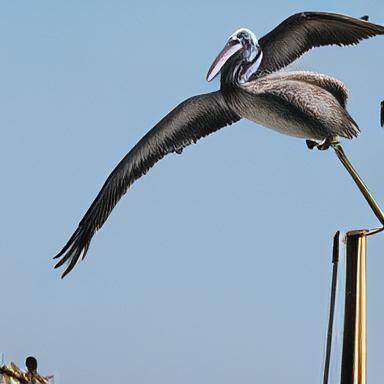} & 
         \includegraphics[trim=2 0 0 2,clip,width=0.15\columnwidth]{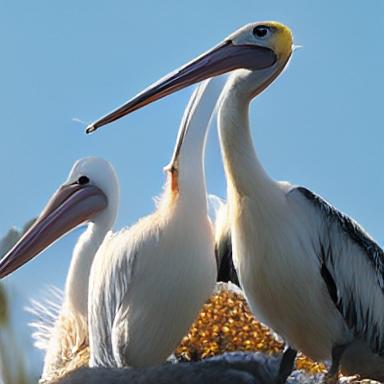} & 
         \includegraphics[trim=2 0 0 2,clip,width=0.15\columnwidth]{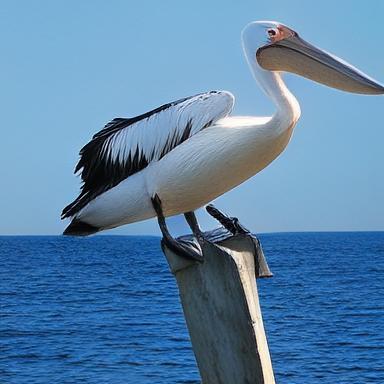} &
         \includegraphics[trim=2 0 0 2,clip,width=0.15\columnwidth]{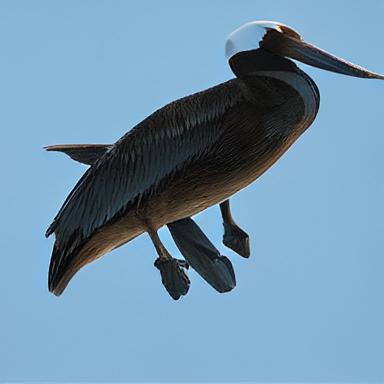} \\

         \includegraphics[trim=2 0 0 2,clip,width=0.15\columnwidth]{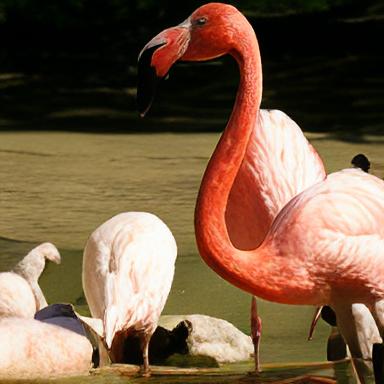} & 
         \includegraphics[trim=2 0 0 2,clip,width=0.15\columnwidth]{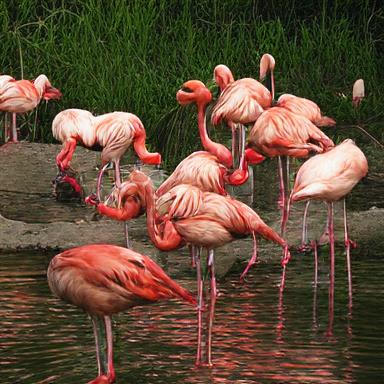} & 
         \includegraphics[trim=2 0 0 2,clip,width=0.15\columnwidth]{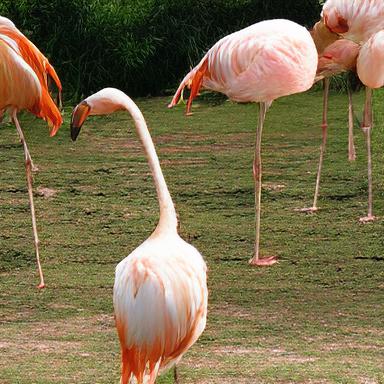} & 
         \includegraphics[trim=2 0 0 2,clip,width=0.15\columnwidth]{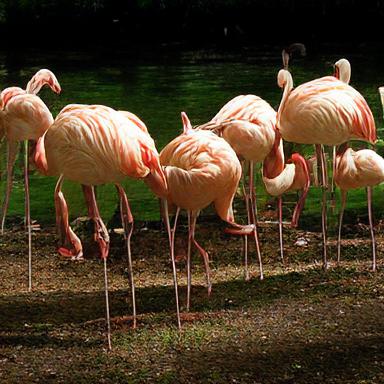} & 
         \includegraphics[trim=2 0 0 2,clip,width=0.15\columnwidth]{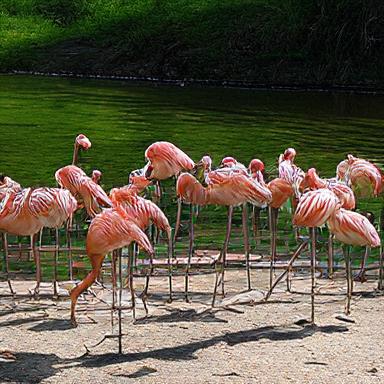} &
         \includegraphics[trim=2 0 0 2,clip,width=0.15\columnwidth]{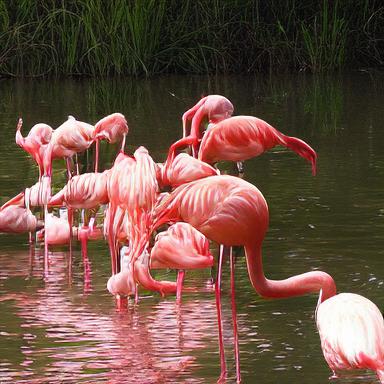} \\

         \includegraphics[trim=2 0 0 2,clip,width=0.15\columnwidth]{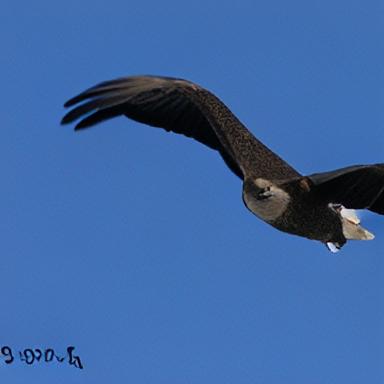} & 
         \includegraphics[trim=2 0 0 2,clip,width=0.15\columnwidth]{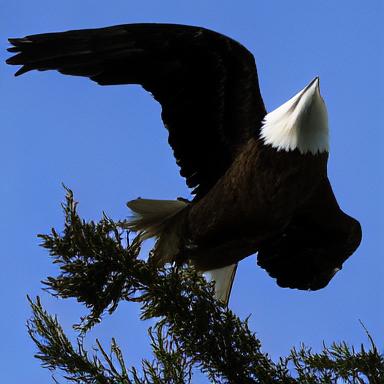} & 
         \includegraphics[trim=2 0 0 2,clip,width=0.15\columnwidth]{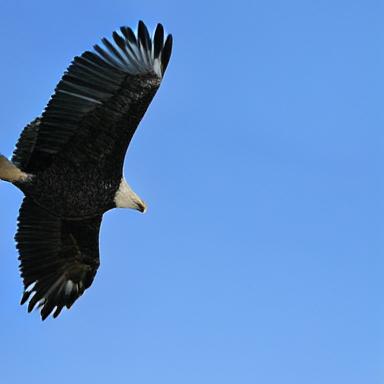} & 
         \includegraphics[trim=2 0 0 2,clip,width=0.15\columnwidth]{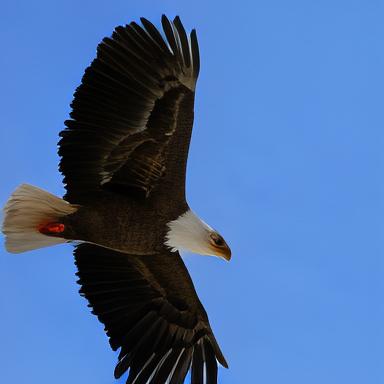} & 
         \includegraphics[trim=2 0 0 2,clip,width=0.15\columnwidth]{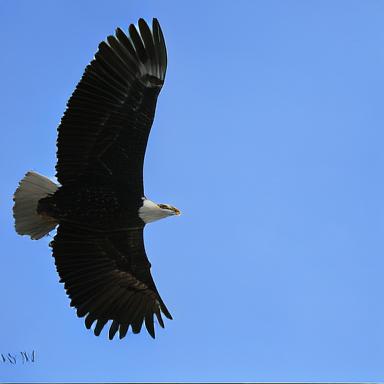} &
         \includegraphics[trim=2 0 0 2,clip,width=0.15\columnwidth]{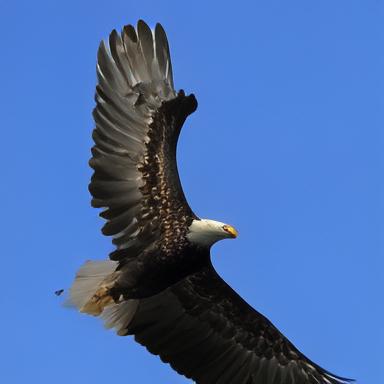} \\

              \includegraphics[trim=2 0 0 2,clip,width=0.15\columnwidth]{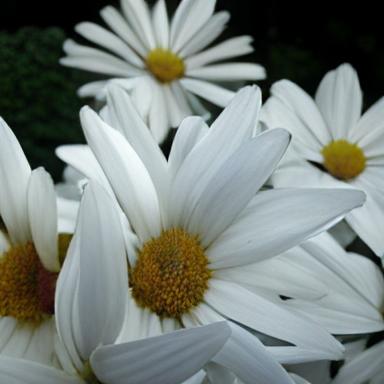} & 
         \includegraphics[trim=2 0 0 2,clip,width=0.15\columnwidth]{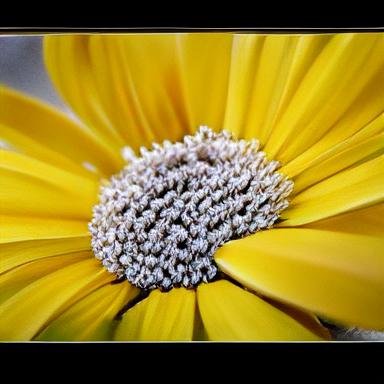} & 
         \includegraphics[trim=2 0 0 2,clip,width=0.15\columnwidth]{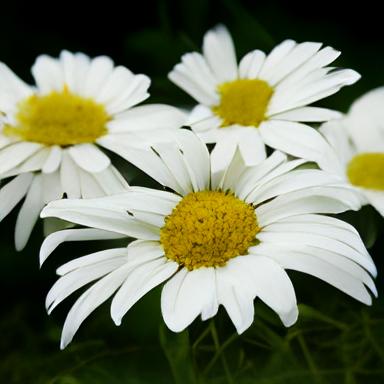} & 
         \includegraphics[trim=2 0 0 2,clip,width=0.15\columnwidth]{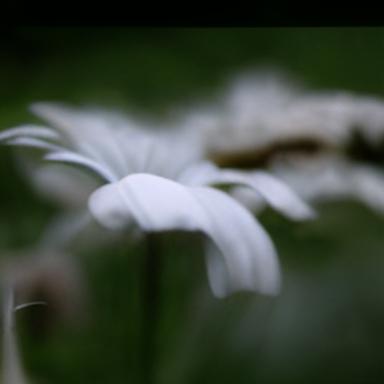} & 
         \includegraphics[trim=2 0 0 2,clip,width=0.15\columnwidth]{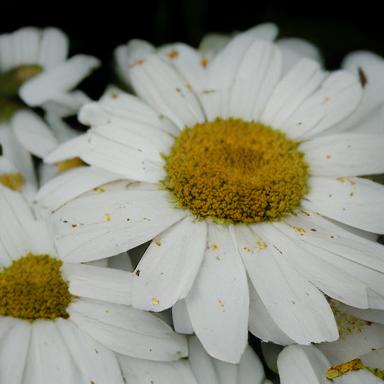} &
         \includegraphics[trim=2 0 0 2,clip,width=0.15\columnwidth]{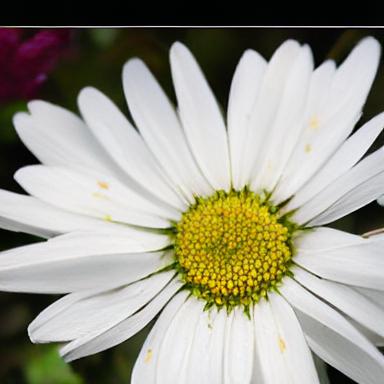} \\

         \includegraphics[trim=2 0 0 2,clip,width=0.15\columnwidth]{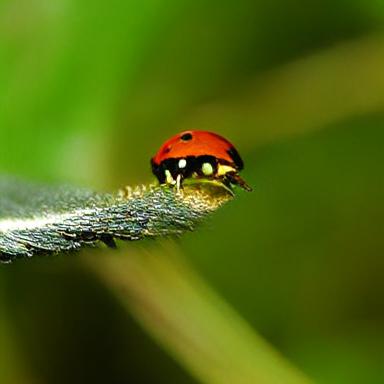} & 
         \includegraphics[trim=2 0 0 2,clip,width=0.15\columnwidth]{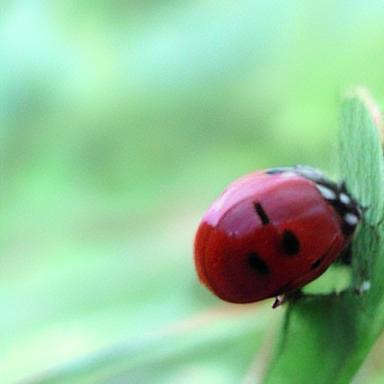} & 
         \includegraphics[trim=2 0 0 2,clip,width=0.15\columnwidth]{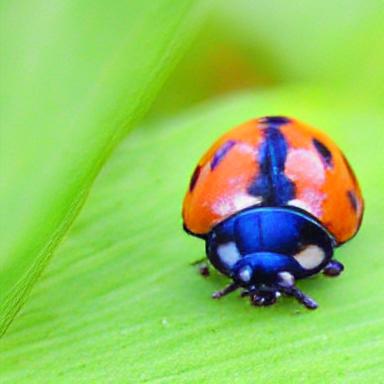} & 
         \includegraphics[trim=2 0 0 2,clip,width=0.15\columnwidth]{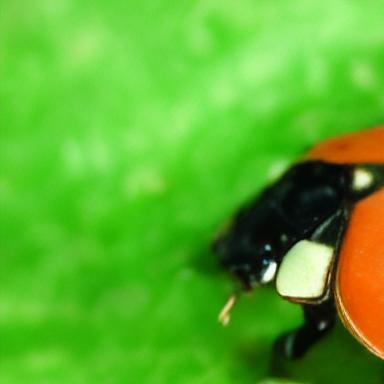} & 
         \includegraphics[trim=2 0 0 2,clip,width=0.15\columnwidth]{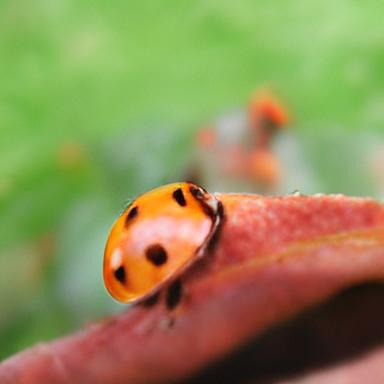} &
         \includegraphics[trim=2 0 0 2,clip,width=0.15\columnwidth]{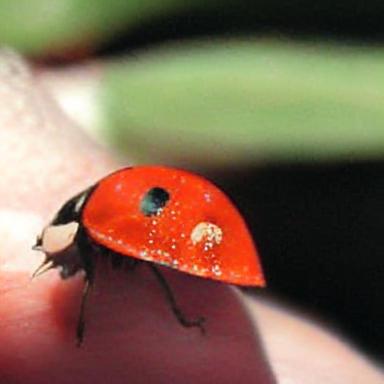} \\
         
    \end{tabular}
    }
    \captionof{figure}{Qualitative comparison for LlamaGen-XL \(\rightarrow\) LlamaGen-L, showing class-conditional samples from the teacher, student, and distilled variants.}
    \label{fig:llamagen_l}
\end{table*}
\begin{table*}[!t]
    \centering
    \def\arraystretch{1.0}
    \resizebox{\textwidth}{!}{
    \setlength\tabcolsep{0pt}
    \footnotesize
    \renewcommand{\arraystretch}{0.5}
    \begin{tabular}{c@{\hskip 0.5mm}c@{\hskip 0.5mm}c@{\hskip 0.5mm}c@{\hskip 0.5mm}c@{\hskip 0.5mm}c}
         Teacher & Student & + KD \cite{hinton2015distilling} & + SeqKD \cite{kim2016sequence} & + GKD \cite{agarwal2024policy} & + \methodname \\
         
         \includegraphics[trim=2 0 0 2,clip,width=0.15\columnwidth]{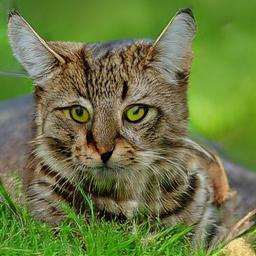} & 
         \includegraphics[trim=2 0 0 2,clip,width=0.15\columnwidth]{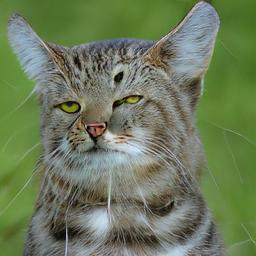} & 
         \includegraphics[trim=2 0 0 2,clip,width=0.15\columnwidth]{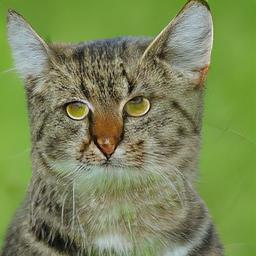} & 
         \includegraphics[trim=2 0 0 2,clip,width=0.15\columnwidth]{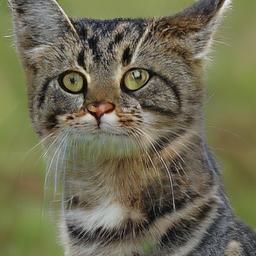} & 
         \includegraphics[trim=2 0 0 2,clip,width=0.15\columnwidth]{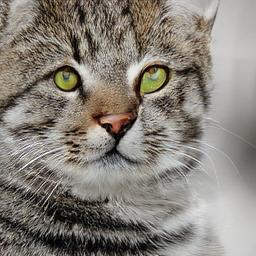} & 
         \includegraphics[trim=2 0 0 2,clip,width=0.15\columnwidth]{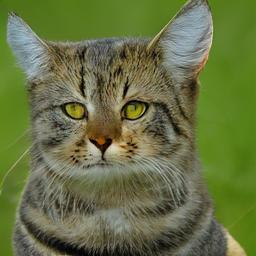}  \\

         \includegraphics[trim=2 0 0 2,clip,width=0.15\columnwidth]{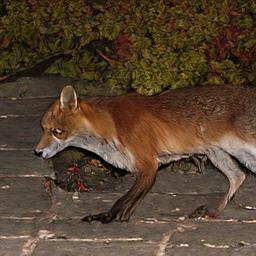} & 
         \includegraphics[trim=2 0 0 2,clip,width=0.15\columnwidth]{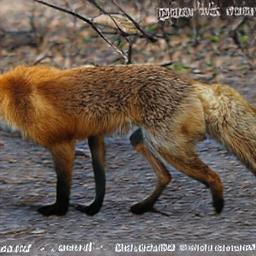} & 
         \includegraphics[trim=2 0 0 2,clip,width=0.15\columnwidth]{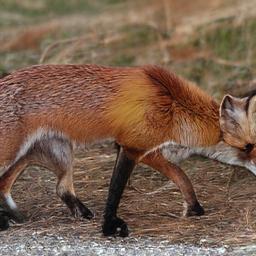} & 
         \includegraphics[trim=2 0 0 2,clip,width=0.15\columnwidth]{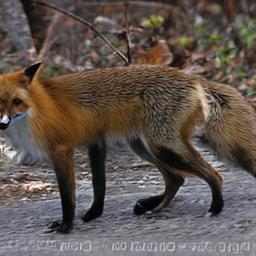} & 
         \includegraphics[trim=2 0 0 2,clip,width=0.15\columnwidth]{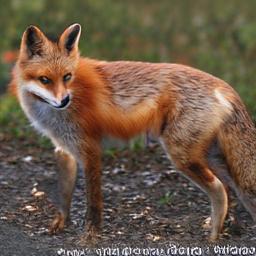} & 
         \includegraphics[trim=2 0 0 2,clip,width=0.15\columnwidth]{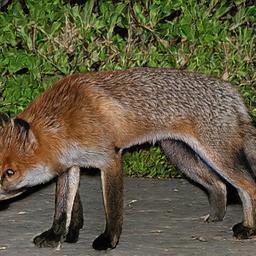}  \\

         \includegraphics[trim=2 0 0 2,clip,width=0.15\columnwidth]{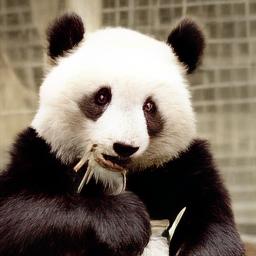} & 
         \includegraphics[trim=2 0 0 2,clip,width=0.15\columnwidth]{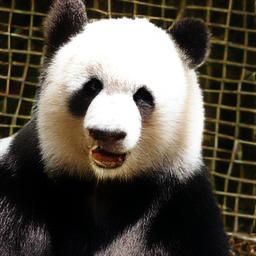} & 
         \includegraphics[trim=2 0 0 2,clip,width=0.15\columnwidth]{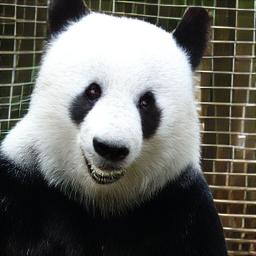} & 
         \includegraphics[trim=2 0 0 2,clip,width=0.15\columnwidth]{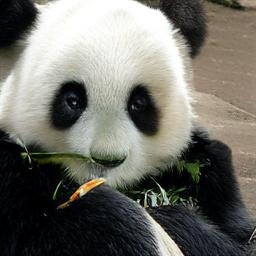} & 
         \includegraphics[trim=2 0 0 2,clip,width=0.15\columnwidth]{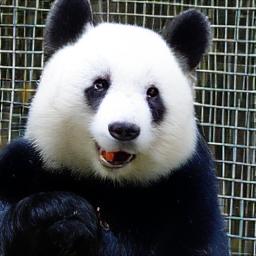} & 
         \includegraphics[trim=2 0 0 2,clip,width=0.15\columnwidth]{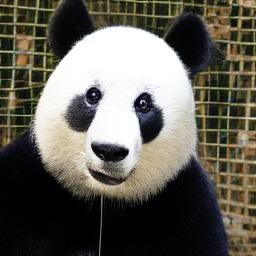}  \\

        \includegraphics[trim=2 0 0 2,clip,width=0.15\columnwidth]{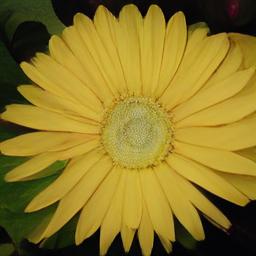} & 
         \includegraphics[trim=2 0 0 2,clip,width=0.15\columnwidth]{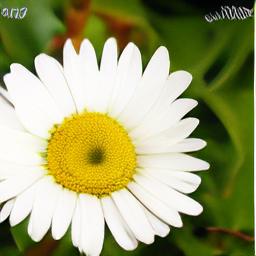} & 
         \includegraphics[trim=2 0 0 2,clip,width=0.15\columnwidth]{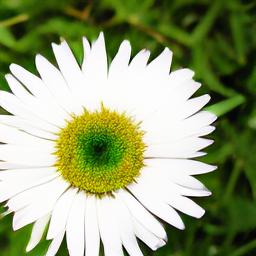} & 
         \includegraphics[trim=2 0 0 2,clip,width=0.15\columnwidth]{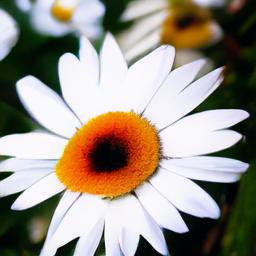} & 
         \includegraphics[trim=2 0 0 2,clip,width=0.15\columnwidth]{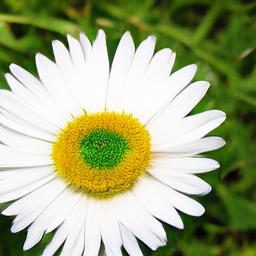} & 
         \includegraphics[trim=2 0 0 2,clip,width=0.15\columnwidth]{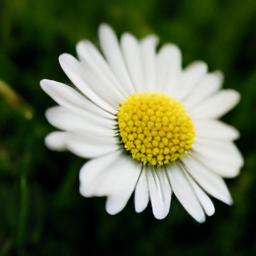}  \\

                  \includegraphics[trim=2 0 0 2,clip,width=0.15\columnwidth]{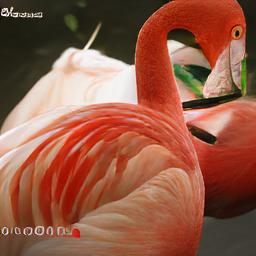} & 
         \includegraphics[trim=2 0 0 2,clip,width=0.15\columnwidth]{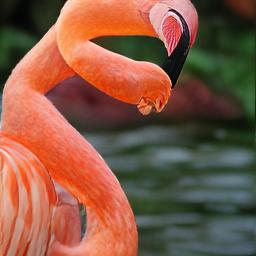} & 
         \includegraphics[trim=2 0 0 2,clip,width=0.15\columnwidth]{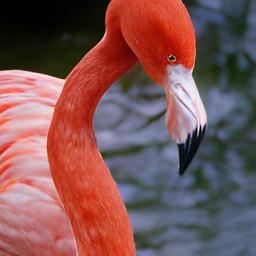} & 
         \includegraphics[trim=2 0 0 2,clip,width=0.15\columnwidth]{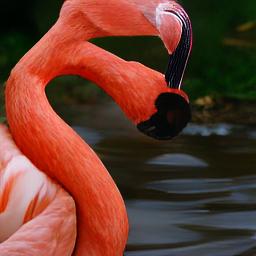} & 
         \includegraphics[trim=2 0 0 2,clip,width=0.15\columnwidth]{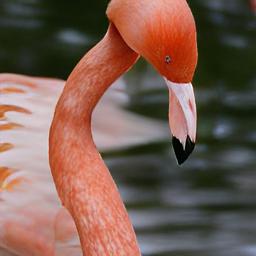} & 
         \includegraphics[trim=2 0 0 2,clip,width=0.15\columnwidth]{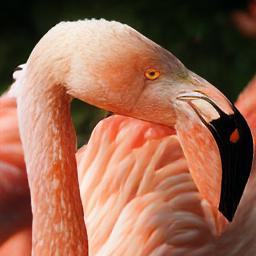}  \\

                           \includegraphics[trim=2 0 0 2,clip,width=0.15\columnwidth]{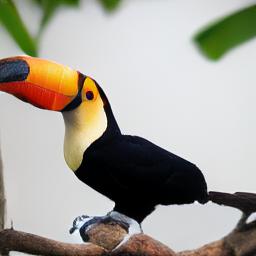} & 
         \includegraphics[trim=2 0 0 2,clip,width=0.15\columnwidth]{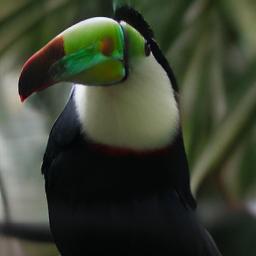} & 
         \includegraphics[trim=2 0 0 2,clip,width=0.15\columnwidth]{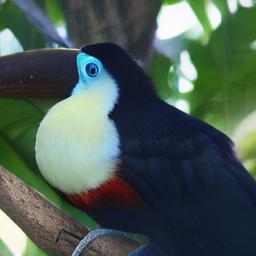} & 
         \includegraphics[trim=2 0 0 2,clip,width=0.15\columnwidth]{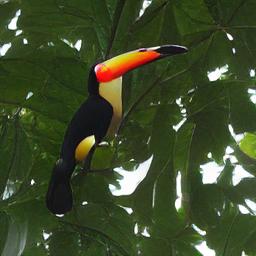} & 
         \includegraphics[trim=2 0 0 2,clip,width=0.15\columnwidth]{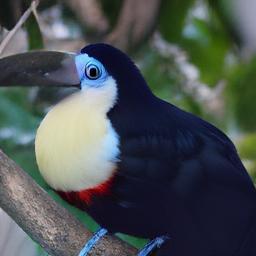} & 
         \includegraphics[trim=2 0 0 2,clip,width=0.15\columnwidth]{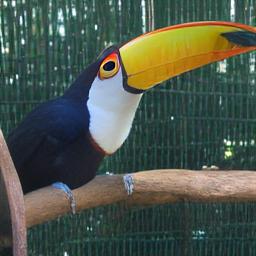}  \\

        \includegraphics[trim=2 0 0 2,clip,width=0.15\columnwidth]{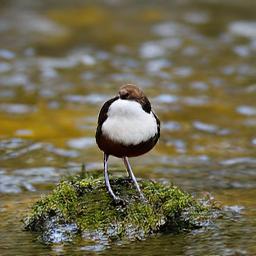} & 
         \includegraphics[trim=2 0 0 2,clip,width=0.15\columnwidth]{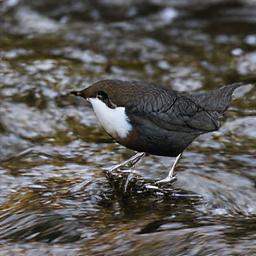} & 
         \includegraphics[trim=2 0 0 2,clip,width=0.15\columnwidth]{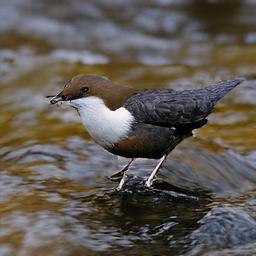} & 
         \includegraphics[trim=2 0 0 2,clip,width=0.15\columnwidth]{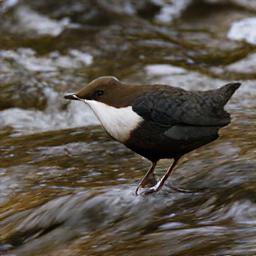} & 
         \includegraphics[trim=2 0 0 2,clip,width=0.15\columnwidth]{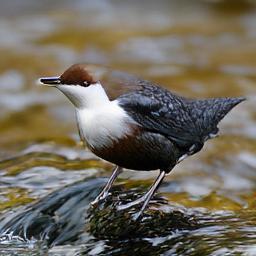} & 
         \includegraphics[trim=2 0 0 2,clip,width=0.15\columnwidth]{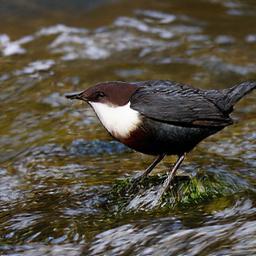}  \\

                \includegraphics[trim=2 0 0 2,clip,width=0.15\columnwidth]{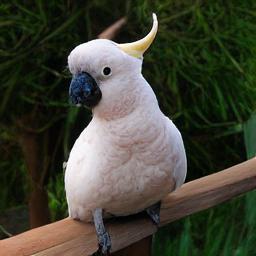} & 
         \includegraphics[trim=2 0 0 2,clip,width=0.15\columnwidth]{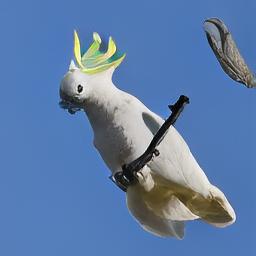} & 
         \includegraphics[trim=2 0 0 2,clip,width=0.15\columnwidth]{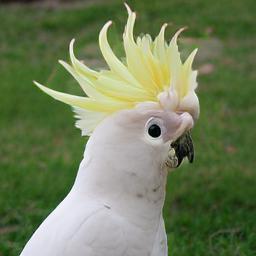} & 
         \includegraphics[trim=2 0 0 2,clip,width=0.15\columnwidth]{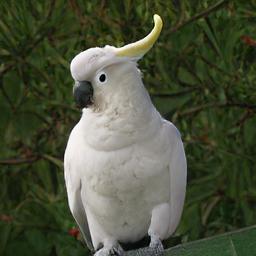} & 
         \includegraphics[trim=2 0 0 2,clip,width=0.15\columnwidth]{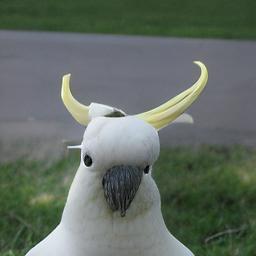} & 
         \includegraphics[trim=2 0 0 2,clip,width=0.15\columnwidth]{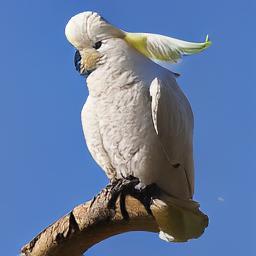}  \\

    \end{tabular}
    }
    \captionof{figure}{Qualitative comparison for ARPG-XL \(\rightarrow\) ARPG-L, showing class-conditional samples from the teacher, student, and distilled variants.}
    \label{fig:arpg_samples}
\end{table*}

\end{document}